%% file: main.tex
\documentclass[10pt,twocolumn,letterpaper]{article}

\usepackage{cvpr}              

\definecolor{cvprblue}{rgb}{0.21,0.49,0.74}
\usepackage[pagebackref,breaklinks,colorlinks,allcolors=cvprblue]{hyperref}

\def\paperID{13971} 
\def\confName{CVPR}
\def\confYear{2025}

\usepackage{multirow}
\usepackage{comment}
\usepackage{wrapfig}
\usepackage{listings}
\usepackage{pdflscape}
\usepackage{rotating}
\usepackage{tcolorbox}
\usepackage{amsmath}
\usepackage{amsfonts}
\usepackage{bm}
\usepackage{amssymb}
\usepackage{makecell}
\usepackage[accsupp]{axessibility}  

\usepackage{graphicx}
\usepackage{booktabs}

\usepackage[accsupp]{axessibility}  

\usepackage{epigraph}

\providecommand\para[1]{\medskip \noindent \textbf{#1}}

\newcommand{\methodname}{{ProsePose }}

\usepackage[font={small}]{caption}
\newcommand{\norm}[1]{\left \lVert #1 \right \rVert}
\DeclareMathOperator*{\argmin}{arg\,min}

\usepackage{microtype}
\usepackage{duckuments} 
\usepackage{cuted}

\usepackage[pagebackref,breaklinks,colorlinks]{hyperref}

\usepackage{orcidlink}

\begin{document}

\title{Pose Priors from Language Models}



\author{$\text{Sanjay Subramanian}^1$
\quad
$\text{Evonne Ng}^1$ 
\quad
$\text{Lea Müller}^1$ 
\quad
$\text{Dan Klein}^1$ \quad
$\text{Shiry Ginosar}^{2,3}$ \quad
$\text{Trevor Darrell}^1$ \\
$\text{}^1$ University of California, Berkeley \\
$\text{}^2$ Google DeepMind \\
$\text{}^3$ Toyota Technological Institute at Chicago \\
\texttt{\{sanjayss,evonne\_ng,mueller,klein,trevordarrell\}@berkeley.edu} \\
\texttt{shiry@google.com}
}



\maketitle

\begin{strip}
    \vspace{-40px}
    \centering
    \includegraphics[width=\linewidth]{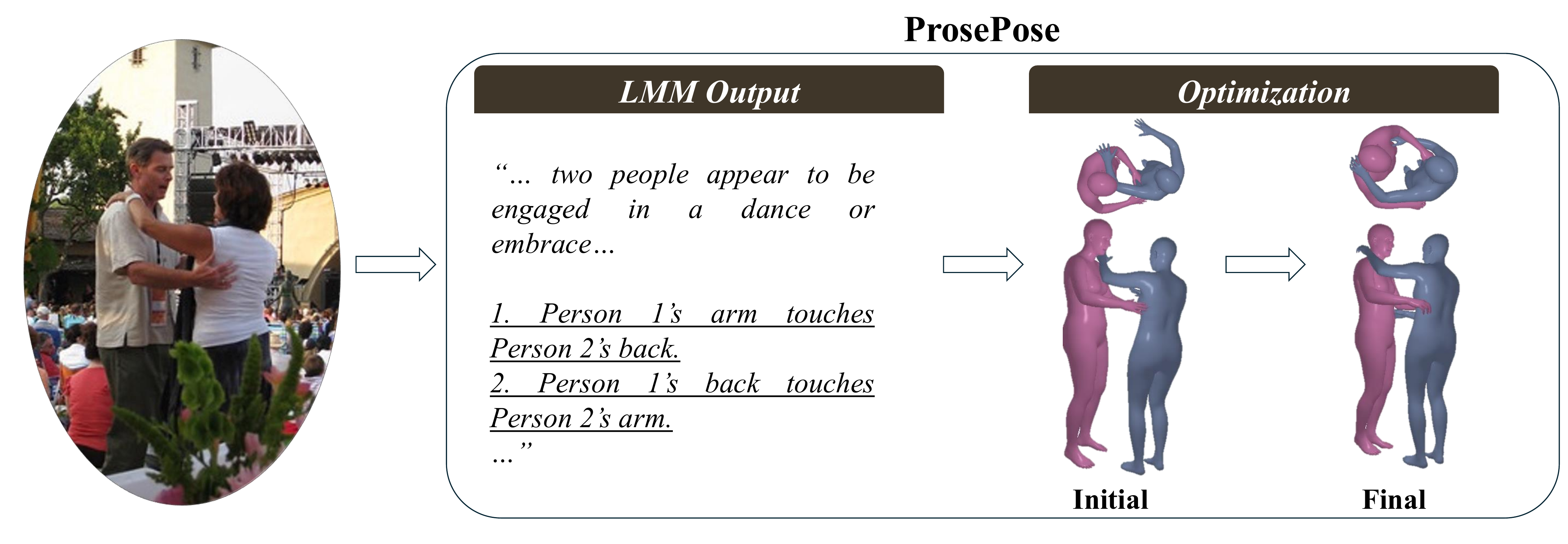}
    \captionof{figure}{\textbf{Optimizing human-to-human contacts in 3D pose.} 
    Our approach leverages the semantic priors of a Large Multimodal Model (LMM) to infer meaningful information about physical contact from images. Instead of relying on human annotations or motion capture data, we extract not only descriptive insights (``... engaged in a dance or embrace ...'') but also structured constraints between body parts (\underline{underlined}). By incorporating these LMM-derived constraints, we refine initial 3D human pose estimates, achieving realistic and semantically consistent reconstructions of contact. This scalable approach opens up new possibilities for contact-aware pose estimation without explicit contact annotations, making it a promising alternative to traditional methods.}
    \label{fig:teaser}
\end{strip}

\begin{abstract}
Language is often used to describe physical interaction, yet most 3D human pose estimation methods overlook this rich source of information. We bridge this gap by leveraging large multimodal models (LMMs) as priors for reconstructing contact poses, offering a scalable alternative to traditional methods that rely on human annotations or motion capture data. Our approach extracts contact-relevant descriptors from an LMM and translates them into tractable losses to constrain 3D human pose optimization. Despite its simplicity, our method produces compelling reconstructions for both two-person interactions and self-contact scenarios, accurately capturing the semantics of physical and social interactions. Our results demonstrate that LMMs can serve as powerful tools for contact prediction and pose estimation, offering an alternative to costly manual human annotations or motion capture data. Our code is publicly available at \url{https://prosepose.github.io}.
\end{abstract}

\input{sections/intro}

\input{sections/related}

\input{sections/method}

\input{sections/results}

\input{sections/conclusion}


%
%
{
    \small
    \bibliographystyle{ieeenat_fullname}
    \bibliography{main}
}
\input{sections/supp.tex}
\end{document}

%% file: sections/intro.tex
\section{Introduction}
Language, as a human artifact, encodes a rich set of social and physical interactions. Over centuries, our vocabulary has evolved to describe the nuances of touch, with words and phrases capturing contexts as varied as hugs, handshakes, or postures in sports and yoga.
Perceiving physical contact is essential for understanding human behavior: e.g. several forms of parent-child contact are associated with affection \cite{takeuchi2010effect,aznar2016parent}, and some forms of self-contact signal stress \citep{harrigan1985self}.

Since written language discusses our physical interactions 
at great length, can large multimodal models (LMMs) trained on images and text correctly perceive physical contact in human pose? This question has practical significance because scenes with contact are challenging for pose estimation methods,
as some body parts are frequently occluded.
This particularly holds for methods solely relying on 2D keypoints which do not convey contact information.
Previously proposed approaches address these issues by curating task-specific datasets via motion capture or human-annotated points of contact between body parts \citep{muller2021self,fieraru2021remips,muller2023generative}. However, collecting these datasets is expensive, and existing publicly available datasets include only tens of thousands of images \citep{fieraru2020three,muller2021self,harmony4d,hi4d}.
If LMMs can accurately identify contact points, they could decrease the cost of curating such datasets.

In this work, we study the efficacy of LMMs as tools for contact prediction in pose estimation. Since LMMs output language rather than pose parameters, answering this question requires a way of eliciting the required information from an LMM and operationalizing it in pose estimation. We introduce a framework, called ProsePose, which prompts an LMM for formatted constraints about physical contact in the image, converts the constraints into a loss function, and optimizes this loss function (jointly with losses from other cues, such as 2D keypoints) to refine initial pose estimates. 

We use \methodname to evaluate LMMs on both 2-person interaction datasets and a dataset of complex yoga poses.
Our framework improves pose estimates compared to strong baselines that do not use contact supervision. In our extensive analysis, we show
that several components are important: mitigating LMM hallucinations by aggregating predictions from several LMM samples, careful construction of the prompt and loss functions, and integrating losses from other cues. Finally, we conduct an extensive analysis of LMM predictions and their role in the optimization results. With respect to LMM failures, we find that identifying the chirality of limbs is a particular challenge for LMMs. While existing supervised methods excel by training on large amounts of supervised data with contact labels, we show that we can extract useful priors for contact prediction from pretrained LMMs \textit{without fine tuning}. 

In summary, our contributions are (1) we introduce a framework for applying LMMs as contact prediction tools in pose estimation, (2) we show that our framework can improve the quality of pose estimates in 2-person and 1-person settings, and (3) we provide an analysis of the components of our framework and LMM failure modes.

%% file: sections/related.tex
\vspace{-1mm}
\section{Related Work}
\noindent\textbf{3D human pose reconstruction.}
Reconstructing 3D human poses from single images is an active area of research. Prior works have explored optimization-based approaches~\citep{SMPL-X:2019, guan2009estimating, lassner2017unite, pavlakos2019expressive, rempe2021humor} or pure regression~\citep{kanazawaHMR18,arnab2019exploiting, guler2019holopose, joo2021exemplar, kolotouros2019learning} to estimate the 3D body pose given a single image. HMR2 \citep{goel2023humans} is a recent state-of-the-art regression model in this line of work.
Building on these 
approaches, some methods have looked into reconstructing multiple individuals jointly from a single image. These methods~\citep{zanfir2018monocular, jiang2020coherent, sun2021monocular} use deep networks to reason about multiple people in a scene to directly output multi-person 3D pose predictions.
BEV~\citep{sun2022putting} accounts for the relative proximity of people explicitly using relative depth annotations to reason about proxemics when 
placing each individual in the scene (e.g.~relative depth of people). However, approaches in both categories generally do not accurately capture physical contact between parts of a single person or between people \citep{muller2023generative,muller2021self}.

\medskip
\noindent\textbf{Contact inference in 3D pose reconstruction.}
3D pose reconstruction is especially challenging when there is self-contact or inter-person contact. This has motivated a line of work on pose reconstruction approaches tailored for these settings.
~\cite{muller2021self} focuses on predicting self contact regions for 3D pose estimation by leveraging a dataset with contact annotations to model complex poses such as 
crossed arms.
~\cite{fieraru2020three} introduces the first dataset with hand-annotated ground-truth contact labels between two people. REMIPS~\citep{fieraru2021remips} and BUDDI~\citep{muller2023generative} train models on the person-to-person contact maps in this data in order to
improve 3D pose estimation of multiple people from a single image. CloseInt~\citep{huang2024closely} trains a physics-guided diffusion model on two-person motion capture data for this task.
However, contact annotations, which are crucial for these approaches, are expensive to acquire. Our method does not require any training on such annotations.
Instead, we leverage an LMM's implicit knowledge of pose to constrain pose optimization to capture both self- and person-to-person contact.

\medskip
\noindent\textbf{Language priors on human pose.}
There exists a plethora of text to 3D human pose and motion datasets~\citep{BABEL:CVPR:2021, Guo_2022_CVPR, Plappert2016}, which have enabled work focused on generating 3D motion sequences of a single person performing a general action~\citep{tevet2023human,jiang2023motiongpt, zhang2023generating}. This line of work has been extended to generating the motion of two people conditioned on text \citep{shafir2023human,liang2023intergen}.

PoseScript~\citep{delmas2022posescript} is 
a method for generating a single person's pose from fine-grained descriptions, which uses training data from motion capture annotated with detailed text. 
PoseFix~\citep{delmas2023posefix} 
introduces a labeled dataset for the task of modifying a pose given a fine-grained description of the desired change and trains a model on this data. 
PoseGPT~\citep{Feng2023PoseGPTCA} is a pose regressor that uses language as part of its training data. However, PoseGPT does not produce better pose estimates than previous state-of-the-art regressors (i.e. regressors that do not use language) and applies only to the one-person setting.
\cite{humanobjectlanguage} uses a text-only LM to improve action-conditioned human-object pose estimation. This method relies on a limited database of action-pose pairs to classify an input pose, and uses an LM to improve pose estimates based on the action retrieved from the database. 

\begin{figure*}[t]
    \centering
    \begin{subfigure}{0.6\textwidth}
        \centering
        \includegraphics[width=\textwidth]{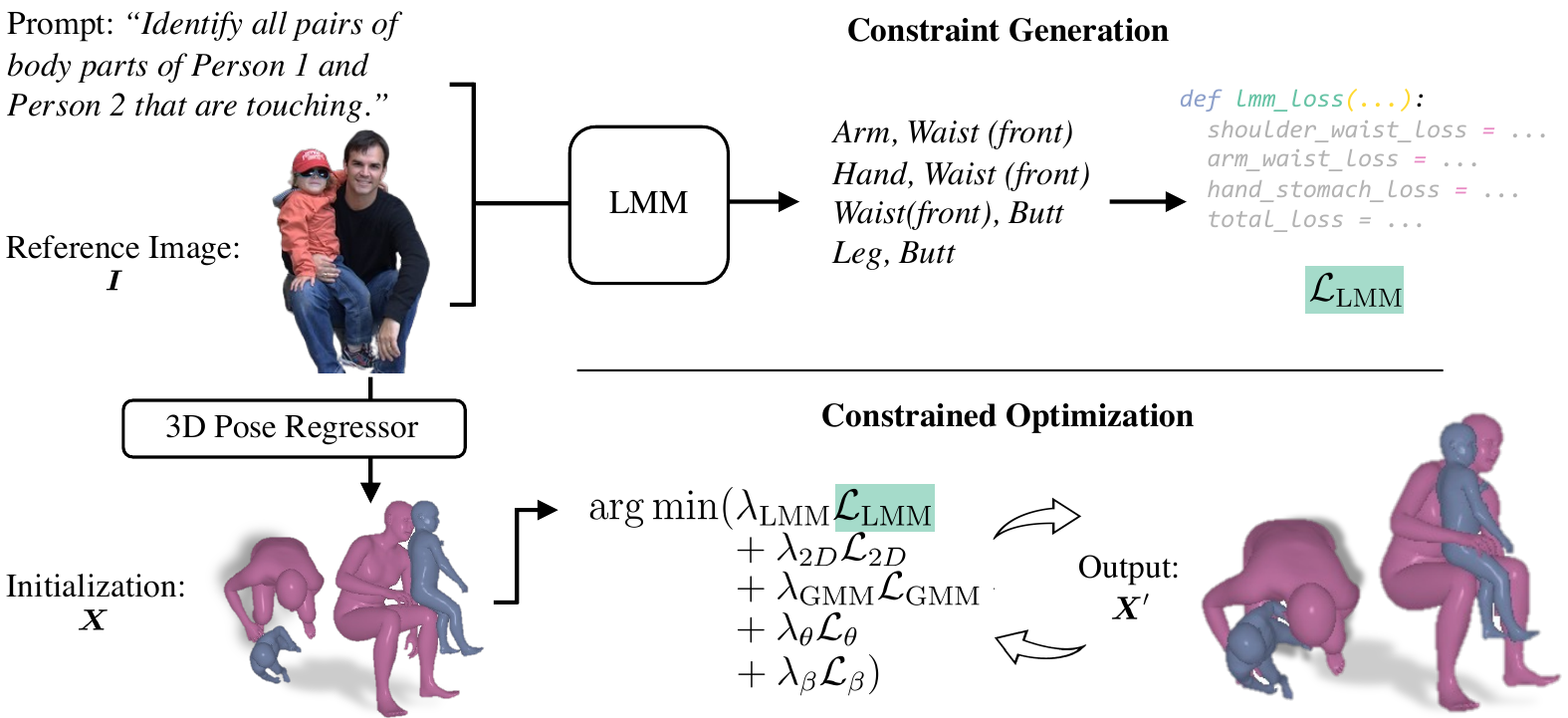}
        \caption{}
        \label{fig:combined-first}
    \end{subfigure}
    \hfill
    \begin{subfigure}{0.3\textwidth}
        \centering
        \includegraphics[width=\linewidth]{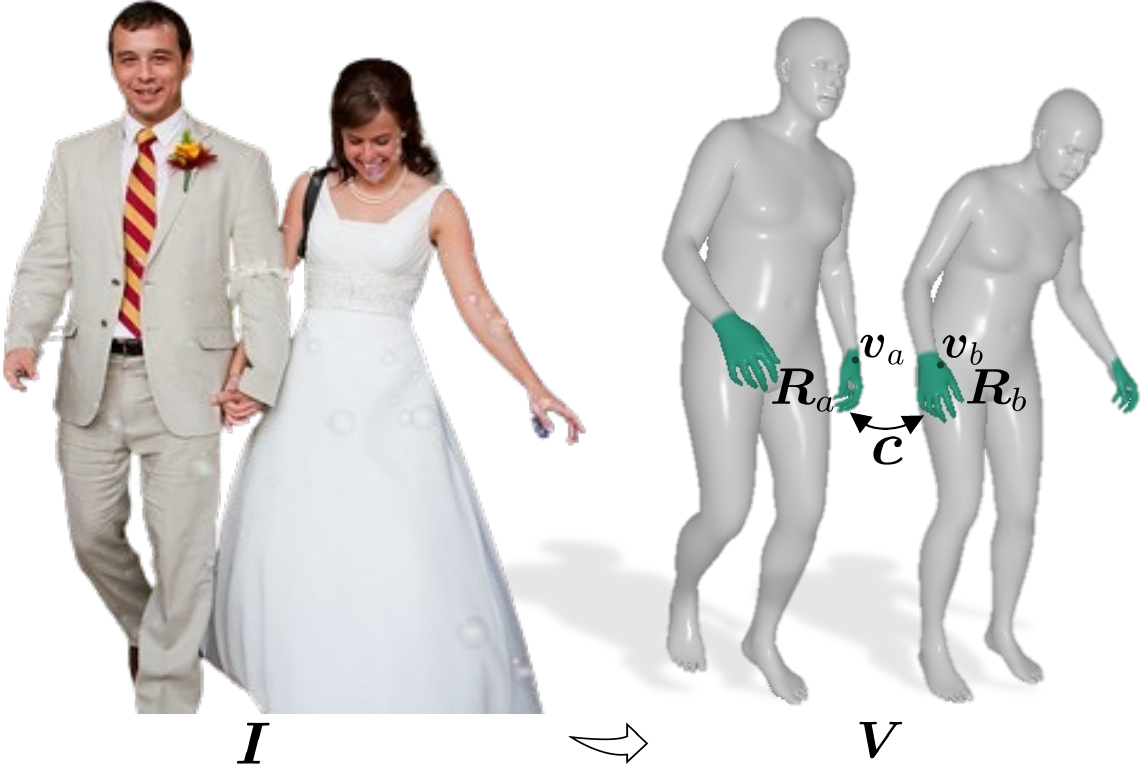}
        \caption{}
        \label{fig:combined-second}
    \end{subfigure}
    \caption{\textbf{LMM-guided Pose Estimation.} \underline{(a) Method overview}: ProsePose takes as input an image of one or two people in contact. We first obtain initial pose estimates for each person from a pose regressor. Then we use an LMM to generate contact constraints, each of which is a pair of body parts that should be touching. This list of contacts is converted into a loss function $\mathcal{L}_{\text{LMM}}$. We optimize the pose estimates using $\mathcal{L}_{\text{LMM}}$ and other losses to produce a refined estimate of each person's pose that respects the predicted contacts. \underline{(b) Defining contact constraints}: Given an image $\boldsymbol{I}$, we can lift each individual into corresponding 3D meshes $\boldsymbol{V}$.
    A contact constraint $\boldsymbol{c}$ is a pair of regions $(\boldsymbol{R}_a, \boldsymbol{R}_b$) in contact. The loss is defined in terms of the distance between the vertices $(\boldsymbol{v}_a, \boldsymbol{v}_b$) on the mesh.}
    \label{fig:combined}
\end{figure*}

Our work differs from previous work on language and pose in several ways. First, whereas all prior work relies on training data with pairs of language and pose, which is expensive to collect, our method leverages the existing knowledge in an LMM to reason about pose from a given image.
Second, prior work in this area focuses on either the one-person or the two-person setting. In contrast, our work 
presents a single framework 
to reason about physical contacts within or between poses. Finally, in scenes with physical contact, we show that our 
method improves the pose estimates of state-of-the-art regressors.

%% file: sections/method.tex
\section{Guiding Pose Optimization with an LMM}

Given an image,
our goal is to estimate the 3D body pose of individuals in the image while capturing the self and cross-person contact points. 
While we cannot trivially use natural language responses (hug, kiss) to directly optimize 3D body poses, we leverage the key insight that LMMs
understand \emph{how} to articulate a given pose (arms around waist, lips touching). We propose a method to structure these articulations into constraints and convert them into loss functions.

More concretely, our framework, illustrated by Figure~\ref{fig:combined-first}, takes as input the image $\boldsymbol{I}$ and the bounding boxes $\boldsymbol{B}$ of the subjects of interest. In the first stage, a pose regressor takes the image and produces a rough estimate of the 3D pose $\boldsymbol{X}^p$ for each individual $p$ in the image. In the second stage, an LMM takes the image and a set of instructions and generates a list of self- or inter-person contact constraints, which we then convert into a loss function (Sec.~\ref{sec:constraint}). Finally, in the third stage, we jointly optimize the generated loss function with several other pre-defined loss terms (Sec.~\ref{sec:optimization}). We refer to our framework as \textbf{ProsePose}.

\subsection{Preliminaries}
\label{sec:preliminaries}
We focus our description on the two-person case to keep the exposition simple. We also demonstrate results on the one-person case, which is simply an extension of the two-person case. In particular, we apply our method to the one-person case by setting $X^0 = X^1$. Please see Appendix~\S~\ref{sec:supp-method}
for details on the differences between the two
cases.

\para{Large Multimodal Models.}
An LMM is a model that takes as input an image and a text prompt and produces text output that answers the prompt based on the image. Our framework is agnostic to the architecture of the LMM. LMMs are typically trained to respond to a wide variety of instructions \citep{liu2023llava,Dai2023InstructBLIPTG}. However, LMMs are prone to hallucination \citep{damonlpsg2023vcd,li2023evaluating}. Handling cases of hallucination is a key challenge when using LMMs. We mitigate this issue by aggregating information across several samples from the LMM.

\para{Pose representation.}
We use a human body model \citep{SMPL-X:2019} to represent each person $p \in \{0,1\}$. The body model is composed of a pose parameter that defines the joint rotations $\boldsymbol{\theta} \in \mathbb{R}^{d_\theta \times 3}$, where $d_\theta$ is the number of joints, and a shape parameter $\boldsymbol{\beta} \in \mathbb{R}^{d_\beta}$, where $d_\beta$ is the dimensions of the shape parameter. We can apply a global rotation $\boldsymbol{\Phi} \in \mathbb{R}^{3}$ and translation $\boldsymbol{t} \in \mathbb{R}^{3}$ to place each person in the world coordinate space. 
The full set of parameters for each person is denoted by $\boldsymbol{X}^p = [\boldsymbol{\theta}^p, \boldsymbol{\beta}^p, \boldsymbol{\Phi}^p, \boldsymbol{t}^p]$.  
For simplicity, we refer to the parameter set $(\boldsymbol{X}^0, \boldsymbol{X}^1)$ as $\boldsymbol{X}$.

These parameters can be plugged into a differentiable function that maps to a mesh consisting of $d_v$ vertices $\boldsymbol{V} \in \mathbb{R}^{d_v \times 3}$. From the mesh, we can obtain
the 3D locations of the body's joints $\boldsymbol{J} \in \mathbb{R}^{d_j \times 3}$. 
From these joints, we can calculate the 2D keypoints $\boldsymbol{K}_{proj}$ by projecting the 3D joints to 2D using the  camera intrinsics $\Pi$ predicted from~\citep{SMPL-X:2019}.
\begin{equation}
    \boldsymbol{K}_{proj} = \Pi\left(\boldsymbol{J}\right) \in \mathbb{R}^{d_j \times 2}.
\end{equation}
\vspace{-0.3cm}
\label{eq:camera}

\para{Vertex regions.}
In order to define contact constraints between body parts, we define a set of \emph{regions} of vertices.  
Prior work on contact has partitioned the body into fine-grained regions~\citep{fieraru2020three}. 
However, since our constraints are specified by a LMM trained on natural language, the referenced body parts are often coarser in granularity. 
We therefore update the set of regions to reflect this language bias by combining these fine-grained regions into larger, more commonly referenced body parts
such as arm, shoulder (front\&back), back, and waist (front\&back). Some regions, e.g. back and waist (back), overlap. Please see Appendix~\S~\ref{sec:supp-regions} 
for a visualization of the coarse regions. Formally, we write $\boldsymbol{R} \in \mathbb{R}^{d_r \times 3}$ to denote a region with $d_r$ vertices, which is part of the full mesh ($\boldsymbol{R} \subset \boldsymbol{V}$).

\para{Constraint definition.}
A contact constraint specifies which body parts from two meshes should touch. 
We define contact constraints as pairs of coarse regions $\boldsymbol{c} = (\boldsymbol{R}_a, \boldsymbol{R}_b)$ between a region $\boldsymbol{R}_a$ of one mesh and $\boldsymbol{R}_b$ of the other mesh, as shown in Figure~\ref{fig:combined-second}. For instance, (``hand'', ``arm'') indicates a hand should touch an arm. 

\subsection{Constraint generation with a LMM}
\label{sec:constraint}
Our key insight is to leverage a LMM to identify regions of contact between different body parts on the human body surface.
As shown in Figure~\ref{fig:combined-first}, we prompt the LMM with an image and ask it to output a list of all 
region pairs that are in contact. 
However, we cannot simply use its output natural language descriptions to directly optimize a 3D mesh.
As such, we 
convert these constraints into a loss function.

\para{LMM-based constraint generation.}
Given the image $\boldsymbol{I}$, we first use the bounding boxes $\boldsymbol{B}$ to crop the part containing the subjects. We then use an image segmentation model to mask any extraneous individuals. 
While cropping and masking the image may remove information, we find the LMMs are relatively robust to missing context, and more importantly, this allows us to indicate which individuals to focus on.
Given the segmented image, we ask the LMM to generate a set $\boldsymbol{C} = \{\boldsymbol{c}_1, ... \boldsymbol{c}_m\}$ of all pairs of body parts that are touching, where $m$ is the total number of constraints the LMM generates for the image.

In the prompt, we specify the full set of coarse regions to pick from.
We find that LMMs fail to reliably reference the left and right limbs correctly or consistently, so the prompt instructs the LMM not to specify chirality for each limb (see Appendix~\ref{sec:supp-quant} for prompt analysis). 
If the LMM uses ``left'' or ``right'' to reference a region, despite an instruction to not do so, we directly use the part of the region with the specified chirality rather than considering both possibilities.

Motivated by the chain-of-thought technique, which has been shown to improve language model performance on reasoning tasks \citep{Wei2022ChainOT}, we ask the LMM to write its reasoning or describe the pose before listing the constraints. For the full prompt used in each setting, please refer to Appendix~\S~\ref{sec:supp-method}. 

We sample $N$ responses from the LMM. 
Below we describe (1) how we convert these natural language responses into 
$N$ sets of constraints $\{\boldsymbol{C}_1, \boldsymbol{C}_2, ..., \boldsymbol{C}_N\}$ and (2) how we 
convert each constraint set $\boldsymbol{C}_j$ 
into a loss. 

\para{Canonicalizing Region Names and Assigning Chirality.} Given the LMM's output, we must map the mentioned region names to our fixed set of coarse regions. 
Since the LMM may deviate from the names in the prompt, we check for some additional names (see Appendix~\ref{sec:supp-regions}). 
We then filter out contact pairs that occur fewer than $f$ times across constraint sets, where $f$ is a hyperparameter.

Next, we assign a chirality (left/right) to each hand/arm/foot/leg/shoulder region in cases when the LMM does not itself specify the chirality. We enumerate all possible assignments of left/right to these regions and take the one resulting in the minimum loss. In the two-person setting, we only consider assignments satisfying a condition designed for the case in which a region type occurs in multiple constraint pairs (see Appendix~\ref{sec:supp-chirality-condition}).

\para{Loss function generation.}  
We compute a loss for each constraint by mapping the relevant regions to sets of vertices and calculating the minimum distance between vertices in the two sets. In particular, we first specify a mapping between each of the coarse region names and the fine-grained regions from \cite{fieraru2020three}. We then use a mapping from fine-grained regions to SMPL-X vertices (provided by \cite{muller2023generative}) to obtain a set of vertices for each coarse region. Then for each contact pair of coarse regions $\boldsymbol{c} = (\boldsymbol{R}_a, \boldsymbol{R}_b)$ in $\boldsymbol{C}_j$, we define $dist(\boldsymbol{c})$ as the minimum distance between the two regions:
\begin{equation}
    dist(\boldsymbol{c}) = \min \norm{ \boldsymbol{v}_a - \boldsymbol{v}_b }_2 \quad \forall \boldsymbol{v}_a \in \boldsymbol{R}_a, \forall \boldsymbol{v}_b \in \boldsymbol{R}_b
\end{equation}
where $\{\boldsymbol{v}_a, \boldsymbol{v}_b\} \in \mathbb{R}^{3}$.
In practice, the number of vertices in each region can be very large. To make this computation tractable, we first take a random sample of vertices from $\boldsymbol{R}_a$ and from $\boldsymbol{R}_b$ before computing distances between pairs of vertices in these samples.

Furthermore, since the ordering of the people in the LMM constraints is unknown (i.e.~does $\boldsymbol{R}_a$ come from the mesh defined by parameter $\boldsymbol{X}^0$ or ~$\boldsymbol{X}^1$), 
we compute the overall loss for both possibilities and take the minimum. We use $\boldsymbol{c}^\top = (\boldsymbol{R}_b, \boldsymbol{R}_a)$ to denote the flipped ordering. We then sum over all constraints in the list $\boldsymbol{C}_j$:
\begin{align}
    {dist}_{\text{sum}}(\boldsymbol{C}_j) = \min\left(\sum_{\boldsymbol{c} \in \boldsymbol{C}_j} dist(\boldsymbol{c}), \sum_{\boldsymbol{c} \in \boldsymbol{C}_j} dist(\boldsymbol{c}^\top)\right)
\label{eq:dist_sum}
\end{align}

\noindent
Each constraint set sampled from the LMM is likely to contain noise or hallucination. To mitigate this issue, we average over all $N$ losses corresponding to each constraint set to obtain the overall LMM loss. This technique is similar to self-consistency \citep{wang2022self}, which is commonly used for code generation. Concretely, the overall LMM loss is defined as
\begin{align}
    \mathcal{L}_{\text{LMM}} = \frac{1}{N}{\sum_{j = 1}^{N} dist_{sum}(\boldsymbol{C}_j)}
\end{align}
If a constraint set $\boldsymbol{C}_j$ is empty (i.e. the LMM does not suggest any contact pairs), then we set $dist_{\text{sum}}(\boldsymbol{C}_j) = 0$. If there are several such constraint sets, we infer that the LMM has low confidence about the contact points (if any) in the image.
Consequently, we set a threshold $t$ and if the number of empty constraint sets is at least 
$t$, we gracefully backoff to the appropriate baseline optimization procedure (described in Sections~\ref{sec:results-two} and~\ref{sec:results-one} for each setting). We also backoff to the baseline if the LMM-based optimization diverges.

\subsection{Constrained pose optimization}
\label{sec:optimization}

Drawing from previous optimization-based approaches~\citep{muller2023generative, bogo2016keep, pavlakos2019expressive}, we employ
several additional losses in the optimization. We then minimize the joint loss to obtain a refined subset of the body model parameters $\boldsymbol{X}' = [\boldsymbol{\theta}', \boldsymbol{\beta}'$, $\boldsymbol{t}']$:
\begin{align}
    [\boldsymbol{\theta}', \boldsymbol{\beta}', \boldsymbol{t}'] = \argmin ( &\lambda_{\text{LMM}}\mathcal{L}_{\text{LMM}} + \lambda_{\text{GMM}} \mathcal{L}_{\text{GMM}} \nonumber
    + \lambda_{\beta} \mathcal{L}_{\beta} \\
    &\;+ \lambda_{\theta} \mathcal{L}_{\theta} \nonumber
    + \lambda_{2D} \mathcal{L}_{2D} + \lambda_{P} \mathcal{L}_{P} )
\label{eq:full}
\end{align}
Following \cite{muller2023generative}, we divide the optimization into two stages. In the first stage, we optimize all three parameters. In the second stage, we optimize only $\boldsymbol{\theta}$ and $\boldsymbol{t}$, keeping the shape $\boldsymbol{\beta}$ fixed.
We detail the remaining losses below.

\para{Pose and shape priors.} We compute a loss $\mathcal{L}_{\text{GMM}}$ based on the Gaussian Mixture pose prior of \cite{bogo2016keep} and a shape loss $\mathcal{L}_{\beta} = \norm{\boldsymbol{\beta}}^2_2$, which penalizes extreme deviations from the body model's mean shape.

\para{Initial pose loss.} To ensure we do not stray too far from the initialization, we penalize large deviations from the initial pose $\mathcal{L}_{\theta} = ||\boldsymbol{\theta}' -\boldsymbol{\theta}||_2^2$.

\para{2D keypoint loss.} Similar to  BUDDI~\citep{muller2023generative}, for each person in the image, we obtain pseudo ground truth 2D keypoints and their confidences from OpenPose~\citep{8765346} and ViTPose~\citep{xu2022vitpose}. 
Given this pseudo ground truth, we merge all the keypoints into $\boldsymbol{K} \in \mathbb{R}^{d_j \times 2}$, and their corresponding confidences into $\boldsymbol{\gamma} \in \mathbb{R}^{d_j}$. 
From the predicted $\boldsymbol{X}'$, we can compute the 2D projection of each 3D joint location using Equation~\ref{eq:camera}. 
Then, the 2D keypoint loss is defined as:
\begin{align}
    \mathcal{L}_{2D} = \sum_{j=1}^{d_j} \boldsymbol{\gamma} (\boldsymbol{K}_{proj}-\boldsymbol{K})^2
\end{align}

\para{Interpenetration loss.} To prevent parts of one mesh from being in the interior of the other, we add an interpenetration loss. 
Generically, given two sets of vertices $\boldsymbol{V}_0$ and $\boldsymbol{V}_1$, we use winding numbers to compute the subset of $\boldsymbol{V}_0$ that intersects $\boldsymbol{V}_1$, which we denote as $\boldsymbol{V}_{0,1}$. Similarly, $\boldsymbol{V}_{1,0}$ is the subset of $\boldsymbol{V}_1$ that intersects $\boldsymbol{V}_0$. The interpenetration loss is then defined as
\begin{align}
    \mathcal{L}_P = \sum_{x \in \boldsymbol{V}_{0,1}} \min_{v_{1} \in \boldsymbol{V}_1} \norm{x-v_1}_2^2 + \sum_{y \in \boldsymbol{V}_{1,0}} \min_{v_0 \in \boldsymbol{V}_0} \norm{y-v_0}_2^2
\end{align}
For efficiency, this loss is computed on low-resolution versions of the two meshes (roughly 1000 vertices per mesh).

%% file: sections/results.tex
\section{Experiments}

\para{Implementation details.}
Following prior work on two-person pose estimation \citep{muller2023generative}, we use BEV~\citep{sun2022putting} to initialize the poses since it was trained to predict both the body pose parameters and the placement of each person in the scene. However, on the single person yoga poses, we find that the pose parameter estimates of HMR2~\citep{goel2023humans} are much higher quality, so we initialize the body pose using HMR2.

\noindent
We use the SMPL-X~\citep{SMPL-X:2019} body model and (unless specified otherwise) GPT4-V \citep{gpt4} as the LMM with temperature $=0.7$ when sampling from it. GPT4-V refers to the \texttt{gpt-4-vision-preview} model in the OpenAI API: \href{https://platform.openai.com}{platform.openai.com}. In the OpenAI API, we use the ``high'' detail setting for image input. Appendix~\ref{sec:supp-quant} provides results using other prompts and other LMMs (LLaVA \citep{liu2024llavanext}, GPT-4o) and a running time analysis. 
Unless otherwise specified, we set $N=20$ samples. 
For all of our 2-person experiments, $f=1$, while $f=10$ in the 1-person setting. We set $t=2$ for the experiment on the CHI3D dataset and $t=N$ for all other experiments. 
The hyperparameters and our prompts were chosen based on experiments on the validation sets.
For other implementation details
refer to Appendix~\S~\ref{sec:supp-method}.

\label{sec:implementation}
\para{Metrics.} As is standard in the pose estimation literature, we report 
Procrustes-aligned Mean Per Joint Position Error (PA-MPJPE) in millimeters. 
This metric finds the best alignment between the estimated and ground-truth pose before computing the joint error.
In the two-person setting, we focus on the \emph{joint} PA-MPJPE, as this evaluation incorporates the relative translation and orientation of the two people.
See Appendix~\S~\ref{sec:supp-quant}
for the per-person PA-MPJPE.

We also include the percentage of correct contact points (PCC) metric introduced by~\citep{muller2023generative}. 
This metric captures the fraction of ground-truth contact pairs that are accurately predicted. For a given radius $r$, a pair is classified as ``in contact" if the two regions are both within the specified radius. 
We use the set of fine-grained regions defined in \cite{fieraru2020three} to compute PCC.
The metric is averaged over $r \in {0, 5, 10, 15, ..., 95}$ mm. 
Since these regions are defined on the SMPL-X mesh topology, we convert the regression baselines-- BEV and HMR2-- from the SMPL mesh topology to SMPL-X to compute this metric. Please see the Appendix~\S~\ref{sec:supp-pcc} 
for more details on the regions and on the mesh conversion.

Finally, we report the F1 of the LMM's predicted coarse region pairs.
Specifically, we compute precision as the proportion of predicted pairs $(\mathbf{R}_a, \mathbf{R}_b)$ such that there is some ground-truth pair $(\mathbf{R}^*_a, \mathbf{R}^*_b)$ for which $\mathbf{R}_a$ overlaps with $\mathbf{R}^*_a$ and $\mathbf{R}_b$ overlaps with $\mathbf{R}^*_b$. We compute recall similarly as a proportion of ground-truth pairs. For datasets that do not provide contact maps, we define the ground-truth pairs from the ground-truth meshes, using a threshold on the minimum distance between regions ($0.01$ and $0.02$ for Hi4D and MOYO, respectively). We ignore chirality when computing these. For a pair of people we take the maximum F1 of the two possible orderings. This metric serves to evaluate the raw output of the LMMs without 3D optimization. 
\begin{table}[t]
\centering 
\setlength{\tabcolsep}{2.5pt} 
\footnotesize
\begin{tabular}{@{}lrrcrrrcrrr@{}}
\toprule 
 & \multicolumn{2}{c}{Hi4D} && \multicolumn{3}{c}{FlickrCI3D} && \multicolumn{3}{c}{CHI3D} \\
& $\text{PM}_{\downarrow}$ & $\text{F1}_{\uparrow}$ && $\text{PM}_{\downarrow}$ & $\text{PCC}_{\uparrow}$ & $\text{F1}_{\uparrow}$ && $\text{PM}_{\downarrow}$ & $\text{PCC}_{\uparrow}$ & $\text{F1}_{\uparrow}$ \\
\cmidrule{2-3} \cmidrule{5-7} \cmidrule{9-11}
\emph{w/o contact sup.} \\
BEV~\citep{sun2022putting} & 144 & -- && 106 & 64.8 & -- && \textbf{96} & 71.4 & -- \\
Heuristic & 116 & -- && 67 & 77.8 & -- && 105 & 74.1 & -- \\
\methodname & \textbf{93} & 24 && \textbf{58} & \textbf{79.9} & 13 && 100 & \textbf{75.8} & 23 \\
\midrule
\emph{w. contact sup.} \\
BUDDI \citep{muller2023generative} & 89 & -- && 66 & 81.9 & -- && \textbf{68} & 78.0 & -- \\
BUDDI+\methodname & \textbf{88} & -- && \textbf{65} & \textbf{83.2} & -- &&  69 & \textbf{78.8} & -- \\
\midrule
\emph{Coarse GT contacts} \\
Oracle & 81 & 100 && 43 & 86 & 100 && 83 & 83.8 & 100 \\
\bottomrule
\end{tabular}
\caption{\textbf{Two-person Results.} Joint PA-MPJPE (abbreviated PM) (lower is better), Avg.~PCC (higher is better), and F1 (higher is better). For FlickrCI3D, PA-MPJPE is computed using the pseudo-ground-truth fits. F1 measures the accuracy of coarse contacts predicted by the LMM, while the other two metrics evaluate the quality of the estimated 3D pose. Last line shows results using ground-truth contact pairs of coarse regions (heuristic is still used as the backoff method when needed). \textbf{Bold} indicates best method without/with contact supervision in each column.}
\label{tab:two-person}
\vspace{-3mm}
\end{table}
\subsection{Two-person Pose Refinement}
\label{sec:results-two}

\para{Datasets.}
We evaluate on three datasets, and our dataset processing largely follows \citep{muller2023generative}.
\textbf{Hi4D} \citep{hi4d} is a motion capture dataset of pairs of people interacting. Each sequence has a subset of frames marked as contact frames, and we take every fifth contact frame. We use the images from a single camera, resulting in 241 images. 
\textbf{Flickr Close Interactions 3D (FlickrCI3D)} \citep{fieraru2020three} is a collection of Flickr images of multiple people in close interaction. The dataset includes manual annotations of the contact maps between pairs of people. \citep{muller2023generative} used these contact maps to create pseudo-ground truth 3D meshes and curated a version of the test set to exclude noisy annotations, which has 1403 images.
\textbf{CHI3D} \citep{fieraru2020three} is a motion capture dataset of pairs of people interacting. We present results on the validation set. 
There are 431 images, distributed across 4 cameras. The images come from 126 video sequences, each of which has a single ``contact frame.''

To develop our method, we experimented on the validation sets of FlickrCI3D and Hi4D, and a sample of the training set from CHI3D. 
For our experiments, we can compute the PCC on FlickrCI3D and CHI3D, which have annotated ground-truth contact maps. 
Following~\cite{muller2023generative}, we exclude from evaluation images where BEV or the keypoint detectors, which are used by the baselines as well, fail to detect one of the subjects in the interaction pair.

\para{Baselines}
We compare our estimated poses to the following:
\begin{itemize}
    \item \textbf{BEV~\citep{sun2022putting}} Multi-person 3D pose estimation method. Uses relative depth to reason about spatial placement of individuals in the scene. \methodname, Heuristic, and BUDDI use BEV to initialize pose estimates.
    \item \textbf{Heuristic} A contact heuristic which includes the auxiliary losses in Section~\ref{sec:optimization} as well as a term that minimizes the minimum distance between the two meshes. 
    Introduced by~\citep{muller2023generative}. We use their hyperparameters for this heuristic. This baseline is also used as the backoff method for \methodname when the number of empty constraint sets is at least the threshold $t$ or when the optimization diverges.
    \item \textbf{BUDDI~\citep{muller2023generative}} 
    This method uses a learned diffusion prior to constrain the optimization. We stress that BUDDI requires a large amount of annotated training data on pairs of interacting bodies, which is not used in our method.
\end{itemize}
\begin{figure}[t]
    \centering
    \includegraphics[width=0.45\textwidth]{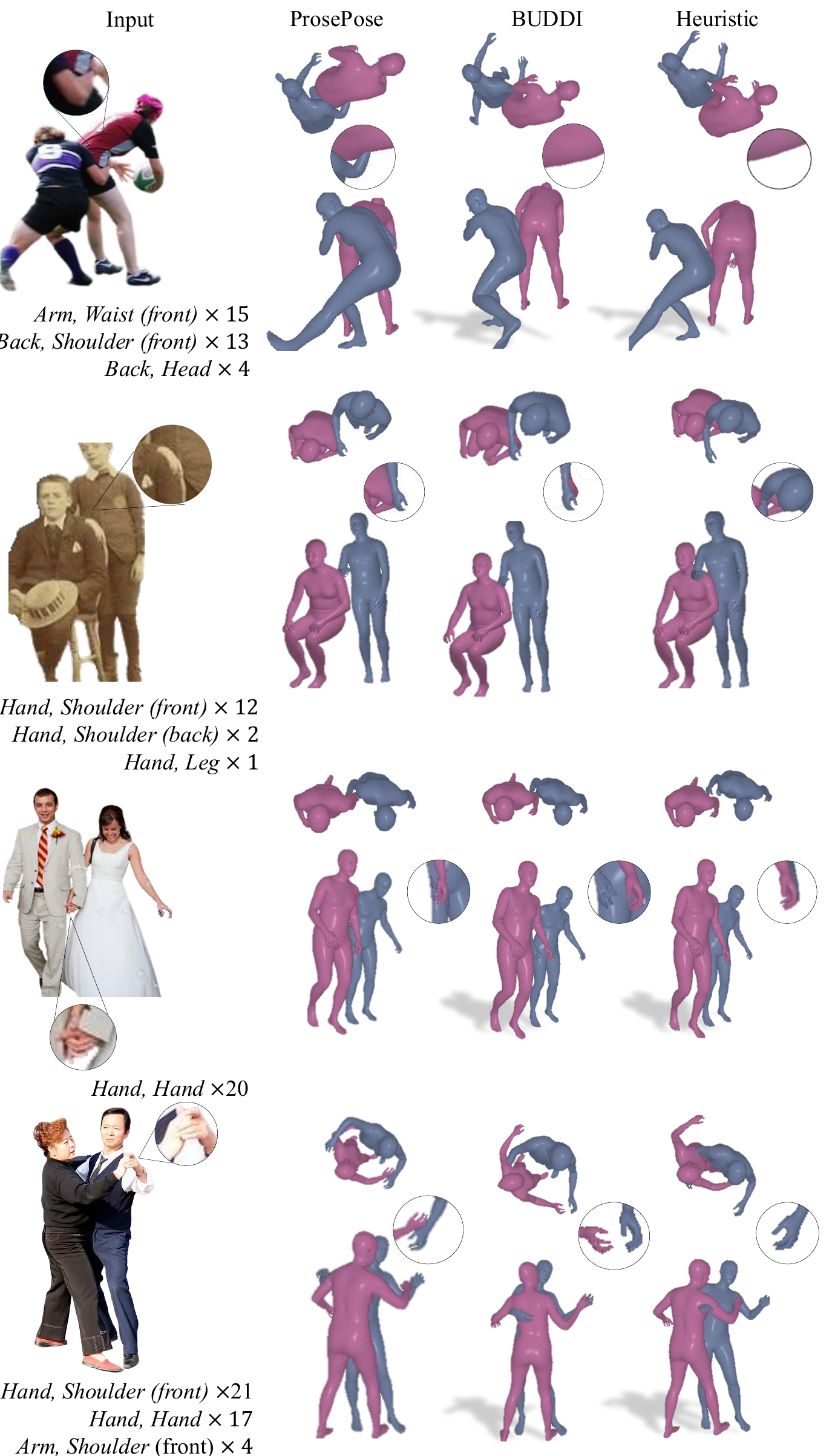}
    \caption{\textbf{Two-person examples} We show qualitative results from \methodname, BUDDI \citep{muller2023generative}, and the contact heuristic. For each example, we show GPT4-V's top 3 constraints and the number of times each constraint was predicted across all 20 samples. Our method correctly reconstructs people in a variety of interactions, and the predicted constraints generally align with each interaction type.}
    \label{fig:results-two}
    \vspace{-3mm}
\end{figure}

\begin{figure}[t]
    \centering
    \includegraphics[width=0.35\textwidth]{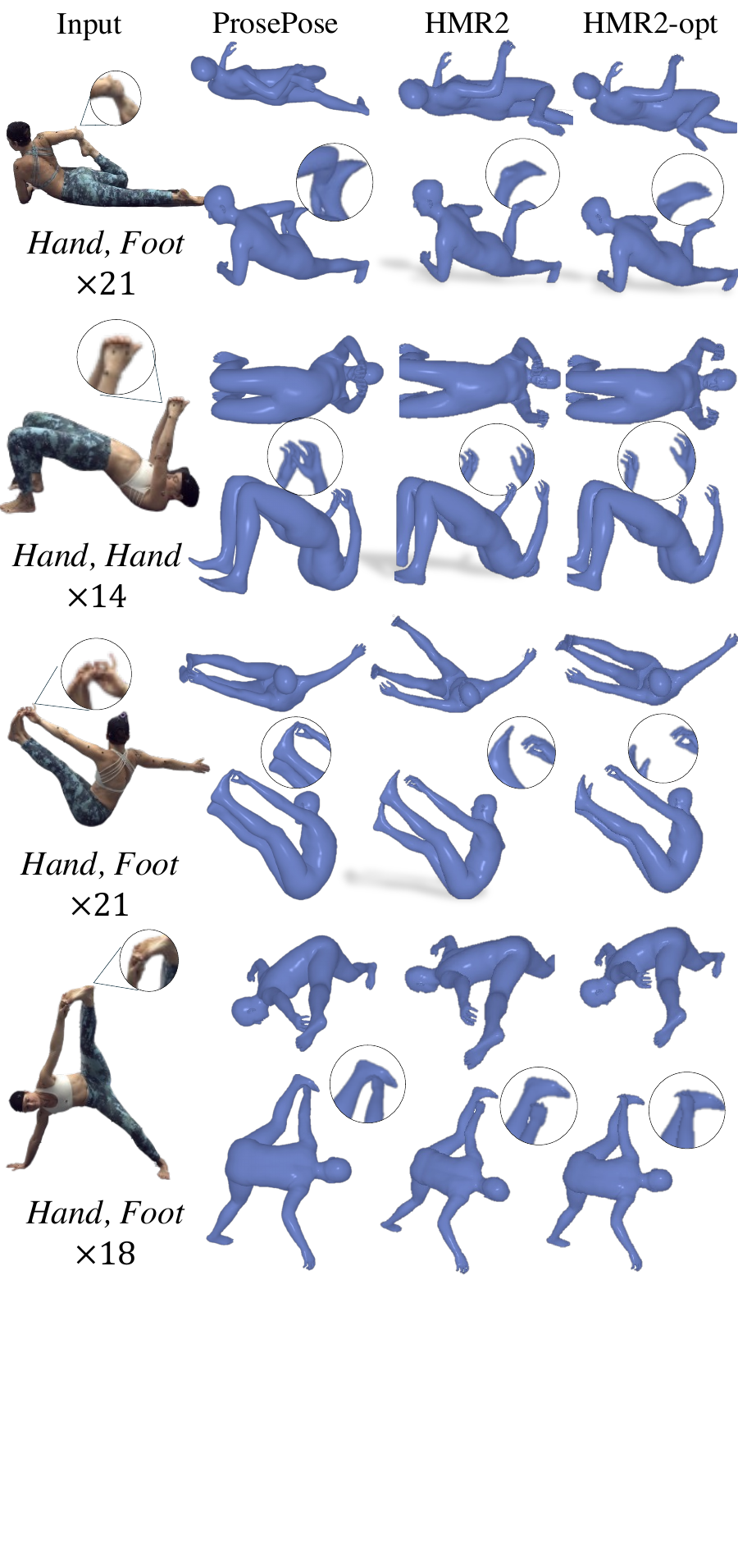}
   \vspace{-2.5cm}
    \caption{\textbf{Single-person examples} We show qualitative results from \methodname, HMR2 \citep{goel2023humans}, and HMR2-optim on complex yoga poses. Each example also shows the constraints that are predicted by the LMM at least $f=10$ times (and are thus used to compute $\mathcal{L}_{\text{LMM}}$) with their counts. \methodname correctly identifies self-contact points and optimizes the poses to respect these contacts.
    }
    \label{fig:yoga}
    \vspace{-3mm}
\end{figure}
\begin{table}[t]\centering \footnotesize

\begin{tabular}{@{}lccccccc@{}}
\toprule 
& \multicolumn{3}{c}{FlickrCI3D $\text{PCC}_{\uparrow}$} && \multicolumn{3}{c}{CHI3D $\text{PCC}_{\uparrow} $}\\
@ $radius [mm]$ & 5 & 10 & 15 && 5 & 10 & 15  \\
\cmidrule{2-4} \cmidrule{6-8}
\emph{W/o contact sup.} \\ 
BEV~\citep{sun2022putting} & 3.6 & 6.3 & 10.8 && 5.8 & 17.4 & 32.5 \\
Heuristic & 14.6 & 33.9 & 49.3 && 11.1 & 28.0 & 45.3 \\
\methodname & \textbf{15.6} & \textbf{39.9} & \textbf{57.1} && \textbf{13.5} & \textbf{35.2} & \textbf{52.5} \\
\midrule
\emph{W/ contact sup.} \\
BUDDI~\citep{muller2023generative} & 18.5 & 44.2 & 61.8 && 15.5 & 39.0 & 56.6 \\
BUDDI+\methodname & \textbf{21.8} & \textbf{49.3} & \textbf{66.4} && \textbf{19.5} & \textbf{43.9} & \textbf{58.8} \\
\bottomrule
\end{tabular}
\caption{\textbf{Two-person PCC.} Percent of correct contact points (PCC) for three different radii $r$ in mm. \textbf{Bold} indicates the best score without/with contact supervision in each column. At the ground-truth contact points, our method brings the meshes closer together than the baselines.}
\label{tab:two-person-pcc}
\vspace{-3mm}
\end{table}

\para{Quantitative Results}
Table~\ref{tab:two-person} provides quantitative results on the three datasets. Across datasets, \methodname consistently improves over the strongest baseline, \textbf{Heuristic}.
On the Hi4D dataset, \methodname reduces 85\% of the gap in PA-MPJPE between \textbf{Heuristic} and the fully supervised \textbf{BUDDI}.  
On the FlickrCI3D and CHI3D datasets, \methodname
narrows the gap in the average PCC between \textbf{Heuristic} and \textbf{BUDDI} by more than one-third.

On CHI3D, \methodname outperforms \textbf{Heuristic} but underperforms \textbf{BEV} in terms of PA-MPJPE.
On the subset of images where we do not default to the heuristic (i.e.~on images where GPT4-V predicts enough non-empty constraint sets), the PA-MPJPE for \methodname and BEV is 86 and 87, respectively.
In other words, in the cases where our method is
actually used, the joint error is slightly less than that of BEV. 
As a result, we can attribute the worse overall error to the poorer performance of the heuristic. The backoff method (which is the heuristic) is used in 13/241 Hi4D examples, 106/1403 Flickr examples, and 224/431 CHI3D examples. Overall, our method improves over the other methods that do not use contact supervision in terms of both joint error and PCC. 
While not the focus of this work, Table~\ref{tab:two-person} also shows that adding \methodname, specifically $\mathcal{L}_{\text{LMM}}$, to BUDDI leads to improved PCC. The F1 scores show that LMMs' raw output is often flawed/incomplete, but our approach mitigates hallucinations in various ways (see the ablation study below). 
The last row in Table~\ref{tab:two-person} shows the performance when the ground-truth coarse contacts (with correct left/right labels) are used in optimization. These results show the benefit of correct coarse contacts and the clear room for improvement from better LMM predictions.

\begin{figure}[ht]
    \includegraphics[trim={0 0 0 2cm},clip,width=\linewidth]{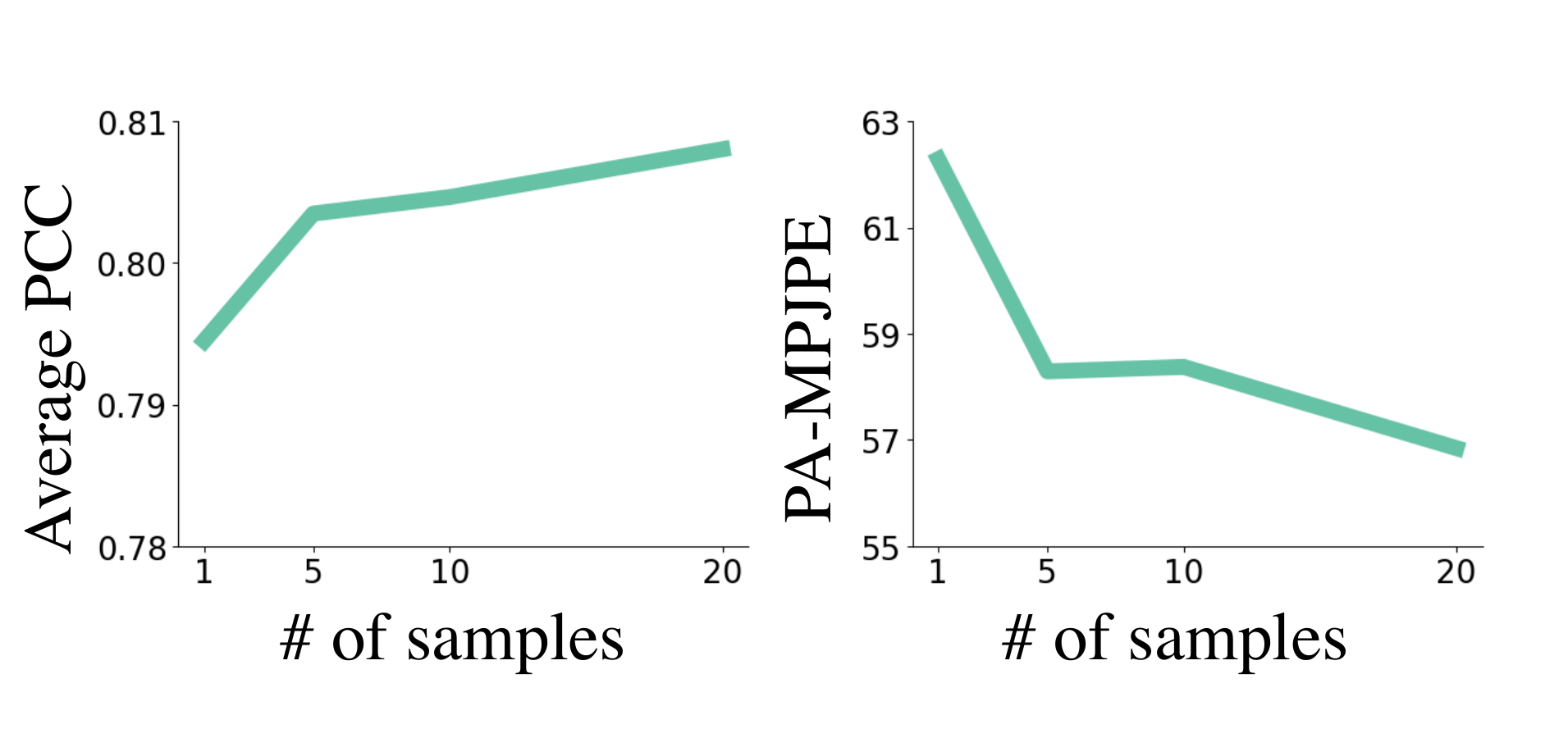}
    \vspace{-7mm}
    \caption{
    \textbf{More samples improve pose estimation.}
    On the FlickrCI3D validation set, taking more samples from the LMM and averaging the resulting loss functions improves joint PA-MPJPE (left) and average PCC (right).}
    \label{fig:sampling}
\end{figure}
Table~\ref{tab:two-person-pcc} shows the PCC for each method at various radii. The results show that \methodname brings the meshes closer together at the correct contact points. On both the FlickrCI3D and CHI3D datasets, \methodname outperforms the other baselines that do not use contact supervision. 
Next, we ablate important aspects of \methodname.
In 
Figure~\ref{fig:sampling}, we show that averaging the loss over several samples from the LMM 
improves performance, mitigating the effect of LMM hallucination. Finally, ablating the various losses in optimization indicates that our LMM-based loss and the 2D keypoint loss have the greatest impact on joint error: Using all losses results in a PA-MPJPE of 81. Removing $\mathcal{L}_{\text{LMM}}$/$\mathcal{L}_{\text{GMM}}$/$\mathcal{L}_{\beta}$/$\mathcal{L}_{\theta}$/$\mathcal{L}_{2D}$/$\mathcal{L}_P$ results in a PA-MPJPE of 138/85/91/84/130/78. 

\para{Qualitative Results} 
Figure~\ref{fig:results-two} shows examples of reconstructions from ProsePose, \textbf{Heuristic}, and \textbf{BUDDI}.
Below each of our predictions, we list the most common constraints predicted by GPT4-V for the image. The predicted constraints correctly capture the semantics of each interaction.
For instance, 
in tango, one person's arm should touch the other's back. In a rugby tackle, a player's arms are usually wrapped around the other player. Using these constraints, \methodname correctly reconstructs a variety of interactions, such as tackling, dancing, and holding hands. In contrast, \textbf{Heuristic} struggles to accurately position individuals and/or predict limb placements, often resulting in awkward distances.

\begin{table}[t]
\centering 
\footnotesize
\setlength{\tabcolsep}{6.5pt}.
\begin{tabular}{@{}lrcccccc@{}}
\toprule 
 & && & \multicolumn{3}{c}{$\text{PCC}_{\uparrow}$ @ $r$} \\
\cmidrule{5-7}
 & \multicolumn{1}{p{0.8cm}}{$\text{PA-MPJPE}_{\downarrow}$} && \multicolumn{1}{p{0.8cm}}{$\text{PCC}_{\uparrow}$} & 5 & 10 & 15 & \multicolumn{1}{c}{$\text{F1}_{\uparrow}$} \\ 
\cmidrule{2-8}
HMR2~\citep{goel2023humans} & 84 && 83.0 & 34.2 & 55.2 & 69.5 & -- \\ 
HMR2+opt & \textbf{81} && 85.2 & 47.7 & 65.5 & 74.6 & -- \\ 
\methodname & 82 && \textbf{87.8} & \textbf{54.2} & \textbf{73.8} & \textbf{81.4} & 25 \\ 
\bottomrule
\end{tabular}
\caption{\textbf{One-person Results.} PA-MPJPE (lower is better) and Avg.~PCC and F1 (higher is better). \methodname captures ground-truth contacts better than the baselines, as shown by the PCC.}
\vspace{-2mm} 
\label{tab:moyo}
\end{table}
\subsection{One-person pose refinement}
\label{sec:results-one}
\para{Datasets}
Next, we evaluate \methodname on a single-person setting.
For this setting, we evaluate on MOYO~\citep{moyo}, a motion capture dataset with videos of a single person performing various yoga poses. 
In total, our test set is composed of 76 examples from a single camera angle (side view). See Appendix~\ref{sec:supp-moyo-details} for further dataset details. Since this dataset does not have annotated region contact pairs, we compute the pesudo-ground-truth contact maps using the Euclidean and geodesic distance following \cite{muller2021self} 
and report PCC for the subset of 67 examples where the ground-truth has self-contact. 

\para{Baselines}
We compare against the following baselines:
\begin{itemize}
    \item \textbf{HMR2}~\citep{goel2023humans} State-of-the-art pose regression method. We use HMR2 to initialize our pose estimates for optimization.
    \item \textbf{HMR2+opt} Optimization procedure that is identical to ours without $\mathcal{L}_\text{LMM}$. It is the default method when the number of empty constraint sets is at least the threshold $t$.
\end{itemize}
Both the quantitative and qualitative results echo the trends discussed in the 2-person setting. 
Table~\ref{tab:moyo} provides the quantitative results. 
The PCC metrics show that our LMM loss improves the predicted self-contact in complex yoga poses relative to the two baselines. The backoff method is used in 43/76 examples.
Figure~\ref{fig:yoga} provides a qualitative comparison of poses predicted by \methodname versus the two baselines. Below each of our predictions, we list the corresponding constraints predicted by GPT4-V. In each case, the predicted constraint captures the correct self-contact, which is reflected in the final pose estimates. 
Using the semantically guided loss, \methodname effectively refines the pose to ensure proper contact between hand-foot or hand-hand, an important detail consistently overlooked by the baselines.
\subsection{Limitations}
\label{sec:limitations}
While \methodname consistently improves contact across settings and datasets, it has limitations which are related to failures of LMMs. First, as shown in Table~\ref{tab:lmm-analysis} (in Appendix), prompting the LMM for left/right labels sometimes leads to worse results, suggesting that LMMs struggle with disambiguating chirality. Improving this approach depends in large part on correctly identifying limbs as left/right.
Another limitation is the use of coarse regions. Future work  could improve by eliciting more fine-grained constraints from an LMM. 
Finally, LMM accuracy varies moderately across camera angles (quantified in Appendix~\ref{sec:supp-quant}).
In Appendix~\S~\ref{sec:supp-failures}, we provide examples of LMM failures.

%% file: sections/conclusion.tex
\section{Conclusion}
We present ProsePose, a framework for refining 3D pose estimates to capture touch accurately using the implicit semantic knowledge of poses in LMMs.
Our key novelty is that we generate structured pose descriptions from LMMs and convert them into loss functions used to optimize the pose.
Our experiments show that in both one-person and two-person settings, \methodname improves over previous baselines that do not use contact supervision.
These results suggest that LMMs may be useful in creating larger datasets with contact annotations, which are otherwise expensive but are crucial for training state-of-the-art priors for pose estimation in situations with physical contact.
More broadly, this work provides evidence that LMMs are promising tools for 3D pose estimation, which likely has implications beyond touch.

\section{Acknowledgements}
\label{sec:ack}
SS, EN, and TD were supported in part by the NSF, DoD, and/or the Berkeley Artificial Intelligence Research (BAIR) industrial alliance program.

%% file: sections/supp.tex
\clearpage
\setcounter{page}{1}
\section*{Appendix for Pose Priors from Language Models}
In this appendix, we provide additional details about our method (Section~\ref{sec:supp-method}), details about metrics (Section~\ref{sec:supp-pcc}), additional quantitative results (Section~\ref{sec:supp-quant}), examples of failure cases (Section~\ref{sec:supp-failures}), experiments with LLaVA 
(Section~\ref{sec:supp-llava}), and more qualitative comparisons (Section~\ref{sec:supp-qual}). We also provide a video overview of the method and qualitative results (\texttt{video.mp4}). 
\section{Additional Method Details}
\label{sec:supp-method}
\subsection{LMM Prompts}
The box below contains our prompt for the two-person experiments.
\begin{tcolorbox}
You are a helpful assistant. You follow all directions correctly and precisely.

For each image, identify all pairs of body parts of Person 1 and Person 2 that are touching.

Write all of these in a Markdown table where the first column is "Person 1 Body Part" and the second column is "Person 2 Body Part".

You can pick which is Person 1 and which is Person 2.

The list of possible body parts is: head, neck, chest, stomach, waist (back), waist (front), back, shoulder (back), shoulder (front), arm, hand, leg, foot, butt.

Do not include left/right.

List ALL pairs you are confident about.

If you are not confident about any pairs, output an empty table.

Carefully write your reasoning first, and then write the Markdown table.
\end{tcolorbox}
The box below contains our prompt for the one-person experiment.
\begin{tcolorbox}
You are a helpful assistant. You answer all questions carefully and correctly.

Identify which body parts of the yogi are touching each other in this image (if any).

Write each pair in a Markdown table with two columns.

Each body part MUST be from this list:

head, back, shoulder, arm, hand, leg, foot, stomach, butt, ground

Do not write "left" or "right".

Describe and name the yoga pose, and then write the Markdown table.

Note that the pose may differ from the standard version, so pay close attention.

Only list a part if you're certain about it.
\end{tcolorbox}
In each setting, the prompt is given as the ``system prompt'' to the GPT-4 API, and the only other message given as input contains the input image with the ``high'' detail setting.

\subsubsection{Ablation Prompts}
\label{sec:supp-ablation-prompts}
Below we give the alternative prompts evaluated in Table~\ref{tab:lmm-analysis}.

The box below contains the prompt that is like the default except that it asks for left/right labels.
\begin{tcolorbox}
You are a helpful assistant. You follow all directions correctly and precisely.
For each image, identify all pairs of body parts of Person 1 and Person 2 that are touching.
Write all of these in a Markdown table where the first column is "Person 1 Body Part" and the second column is "Person 2 Body Part".
You can pick which is Person 1 and which is Person 2.
The list of possible body parts is: head, neck, chest, stomach, waist (back), waist (front), back, shoulder (back), shoulder (front), arm, hand, leg, foot, butt.
For arm/hand/leg/foot/shoulder, prepend "left" or "right".
List each body part separately (don't use plural).

List ALL pairs you are confident about.
If you are not certain about any pairs, output an empty table.
Carefully write your reasoning first, and then write the Markdown table.
\end{tcolorbox}
\begin{figure}
    \includegraphics[width=0.4\textwidth]{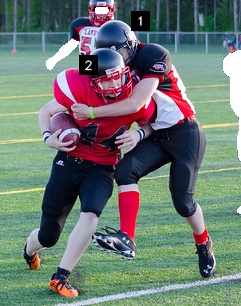}
    \caption{An image with labels for Person 1 and Person 2, from the FlickrCI3D validation set}
    \label{fig:labeledp1p2}
\end{figure}
The box below contains the prompt that is like the default except that the image is labeled with Person 1 and Person 2 and the prompt asks for left/right labels. Figure~\ref{fig:labeledp1p2} shows an example of an input image with labels for Person 1 and Person 2.
\begin{tcolorbox}
You are a helpful assistant. You follow all directions correctly and precisely.
For each image, identify all pairs of body parts of Person 1 and Person 2 that are touching.
Write all of these in a Markdown table where the first column is "Person 1 Body Part" and the second column is "Person 2 Body Part".
The list of possible body parts is: head, neck, chest, stomach, waist (back), waist (front), back, shoulder (back), shoulder (front), arm, hand, leg, foot, butt.
For arm/hand/leg/foot/shoulder, prepend "left" or "right".
List each body part separately (don't use plural).

Only list pairs you are absolutely certain about.
If you are not certain about any pairs, output an empty table.
Carefully write your reasoning first, and then write the Markdown table.
\end{tcolorbox}
The box below contains the prompt for obtaining a pose caption about a pair of people.
\begin{tcolorbox}
Describe the pose of the two people.
\end{tcolorbox}
The box below contains the prompt for rewriting the caption so that it does not contain references to ``left'' and ``right''.
\begin{tcolorbox}
Rewrite the caption below so that it doesn't mention "left" or "right" to describe any hand, arm, foot, or leg. The revised caption should otherwise be identical.
Write only the revised caption and no other text.
\end{tcolorbox}
The box below contains the prompt for converting a caption into a table.
\begin{tcolorbox}
You are a helpful assistant. You will follow ALL rules and directions entirely and precisely.

Given a description of Person 1 and Person 2 who are physically in contact with each other, create a Markdown table with the columns "Person 1 Body Part" and "Person 2 Body Part", listing the body parts of the two people that are guaranteed to be in contact with each other, from the following list. ALL body parts that you list must be from this list. You can choose which person is Person 1 and which is Person 2.
Body parts: "chest", "stomach", "waist (front)", "waist (back)", "shoulder (front)", "shoulder (back)", "back", "hand", "arm", "foot", "leg", "head", "neck", "butt"
Note that "back" includes the entire area of the back.

Include all contact points that are directly implied by the description, not just those that are explicitly mentioned.
If there are no contact points between these body parts that the description implicitly or explicitly implies, your table should contain only the column names and no other rows.

First, write your reasoning. Then write the Markdown table.
\end{tcolorbox}
For the last two prompts above, which do not involve image input, we use the \texttt{gpt-4-0125-preview} version of GPT4 via the OpenAI API.

\subsection{Coarse Regions}
\label{sec:supp-regions}
\begin{figure*}
    \includegraphics[width=\textwidth]{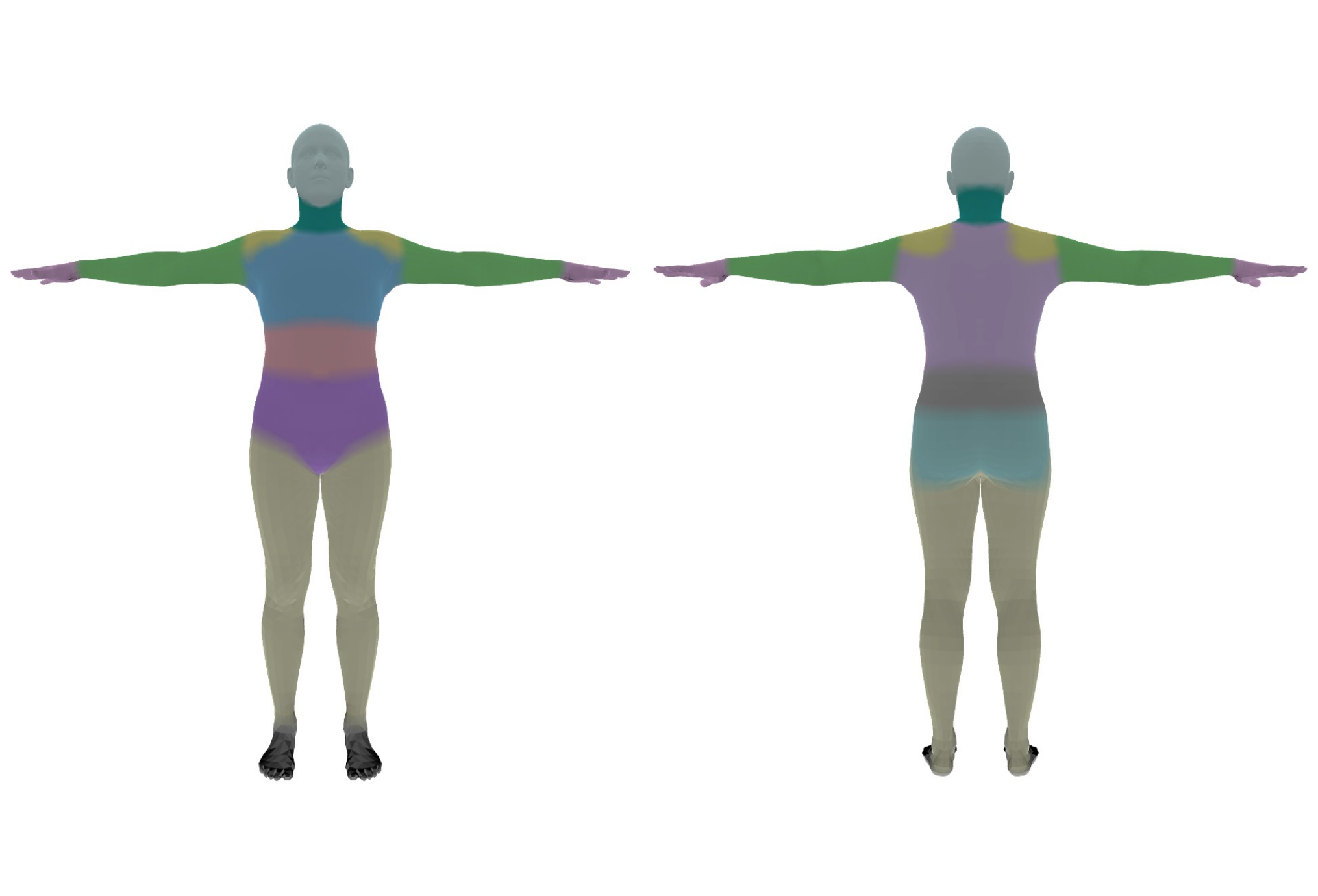}
    \caption{Color-coded coarse regions in the two-person prompt: head, neck, chest, stomach, waist (back), waist (front), back, shoulder (back), shoulder (front), arm, hand, leg, foot, butt. Note that some of these regions overlap. For instance, the ``back'' includes the ``waist (back)'' and ``shoulder (back)'' regions as a subset.}
    \label{fig:coarse-regions-two-person}
\end{figure*}
\begin{figure*}
    \includegraphics[width=\textwidth]{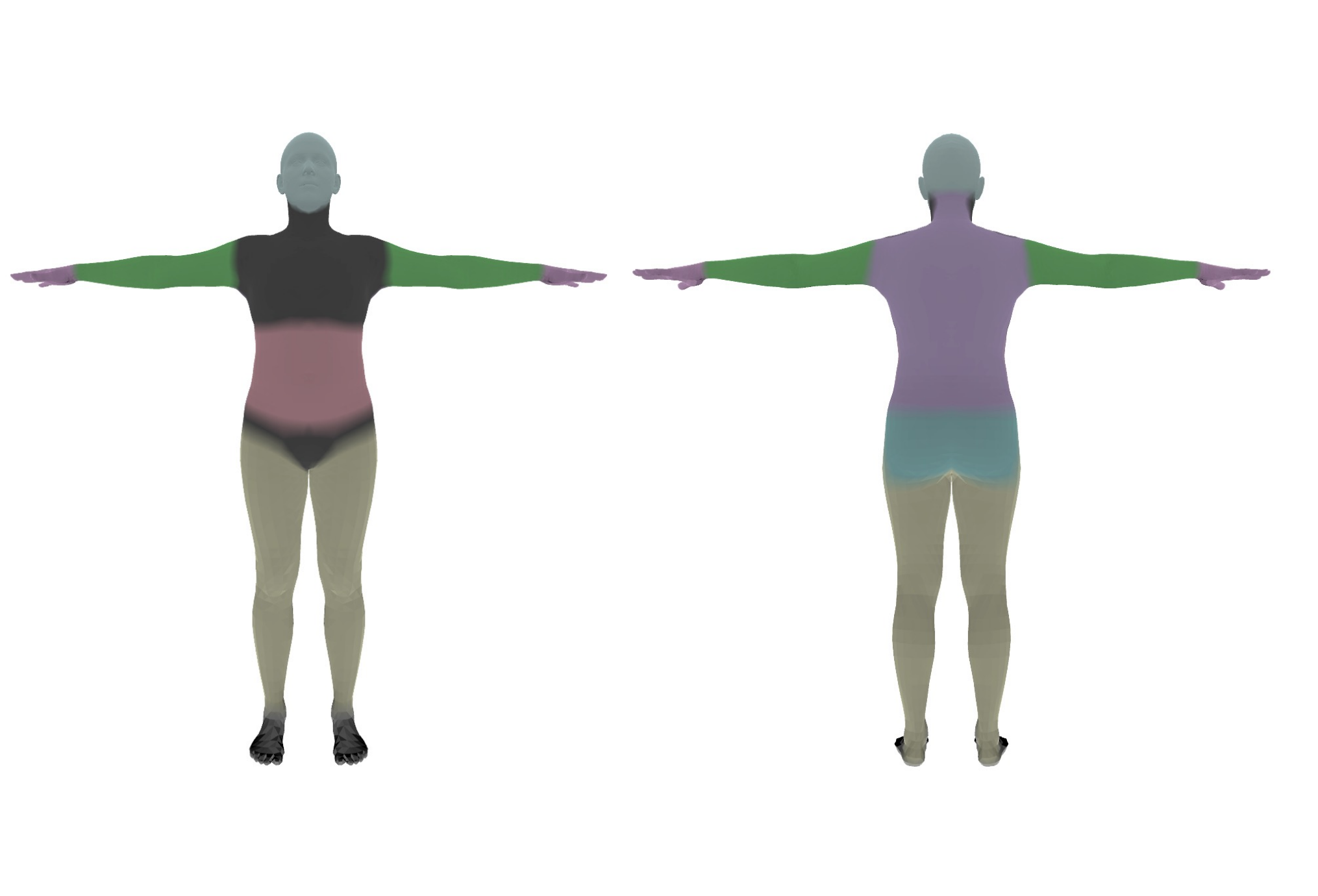}
    \caption{Color-coded coarse regions in the one-person prompt: head, stomach, back, shoulder, arm, hand, leg, foot, butt. Note that the ``chest,'' ``neck,'' and ``waist (front)'' regions are not covered by the regions in the prompt, since they tend to have less importance for contacts in yoga poses.}
    \label{fig:coarse-regions-one-person}
\end{figure*}
Figure~\ref{fig:coarse-regions-two-person} illustrates the coarse regions referenced in the prompt in our two-person experiments. Figure~\ref{fig:coarse-regions-one-person} illustrates the coarse regions referenced in the prompt in our one-person experiments. In the one-person case, the prompt does not mention the ``chest,'' ``neck,'' or ``waist'' regions, since they tend to be less important for contacts in yoga poses, and the front/back shoulders are merged into one region, since the distinction tends to be less important for contacts in yoga poses.

As stated in \S~\ref{sec:constraint}, the procedure converting LMM outputs to loss functions checks for region names other than those listed in the prompt. In particular, it checks for ``waist'' and the left/right variants of ``hand''/``arm''/``foot''/``leg''/``shoulder''/``shoulder (front)''/``shoulder (back)''. 
``waist'' corresponds to the union of ``waist (front)'' and ``waist (back).'' and ``shoulder'' corresponds to the union of ``shoulder (front)'' and shoulder (back).'' 
In the one-person setting, the prompt also specifies ``ground'' as one of the regions, but this is not used in creating loss functions. 

\subsection{Chirality Condition}
\label{sec:supp-chirality-condition}
As mentioned briefly in \S~\ref{sec:constraint}, when enumerating possible chiralities for each constraint, we enforce the following condition in the two-person setting: if the same body part (e.g. ``hand'') is mentioned in at least two separate rows of the table output by the LMM (without any ``left'' or ``right'' prefix) or is mentioned in the plural form (e.g. ``hands''), we enforce that both the left and right limbs of this type must participate in the loss. The motivation for this condition is that when the same constraint applies to both limbs of a given type (\eg ``right hand, back'' and ``left hand, back'') and the chirality is not specified in the constraint set, the two constraints will appear to be the same (\eg ``hand, back''). But often the LMM will list the constraint twice, since there are two different contact points, which activates this condition.

\subsection{Bounding Boxes and Cropping}
\label{sec:bounding-boxes}
As stated in Section 3 of the main paper, we take bounding boxes of the subjects of interest as input and use them to crop the image in order to isolate the person/people of interest when prompting the LMM. For FlickrCI3D, we use the ground-truth bounding boxes of the two subjects of interest. For the other datasets, we use keypoints detected by ViTPose/OpenPose to create the bounding boxes. We use Segment Anything~\citep{kirillov2023segany} as the segmentation model, used to remove extraneous people in the image (we only apply this step for FlickrCI3D, since other datasets are from motion capture). For the single-person MOYO dataset, we manually check that the bounding boxes from the keypoints and the selected HMR2 outputs correspond to the correct person in the image. We note that the baseline HMR2+opt also benefits from this manual checking, since HMR2+opt also depends on the HMR2 outputs and accurate keypoints.

\subsection{Loss Coefficients and Optimization Details}
\label{sec:supp-loss-opt-details}
We set $\lambda_{\text{LMM}} = 1000$ in the 2-person experiments, and $\lambda_{\text{LMM}} = 10000$ in the 1-person setting. In the two-person case, all other loss coefficients are taken directly from \cite{muller2023generative}. In the one-person case, we find that removing the GMM pose prior and doubling the weight on the initial pose loss improves optimization dramatically, likely because the complex yoga poses are out of distribution for the GMM prior. These hyperparameters and our prompts were chosen based on experiments on the validation sets. Furthermore, following~\cite{muller2023generative}, we run both optimization stages for at most 1000 steps. We use the Adam optimizer \citep{Kingma2014AdamAM} with learning rate 0.01.

\subsection{MOYO Dataset Processing Details}
\label{sec:supp-moyo-details}
The dataset provides views from multiple different cameras. We pick a single camera that shows the side view for evaluation.
For each video, we take single frame from the middle as it generally shows the main pose. There is no official test set, and the official validation set consists of only 16 poses. Therefore, we created our own split by picking 79 arbitrary examples from the training set to form our validation set. 
We then combine the remaining examples in the training set with the official validation set to form our test set. 

\section{Experiments}
\subsection{PCC Calculation}
\label{sec:supp-pcc}
\begin{figure*}
    \includegraphics[width=\textwidth]{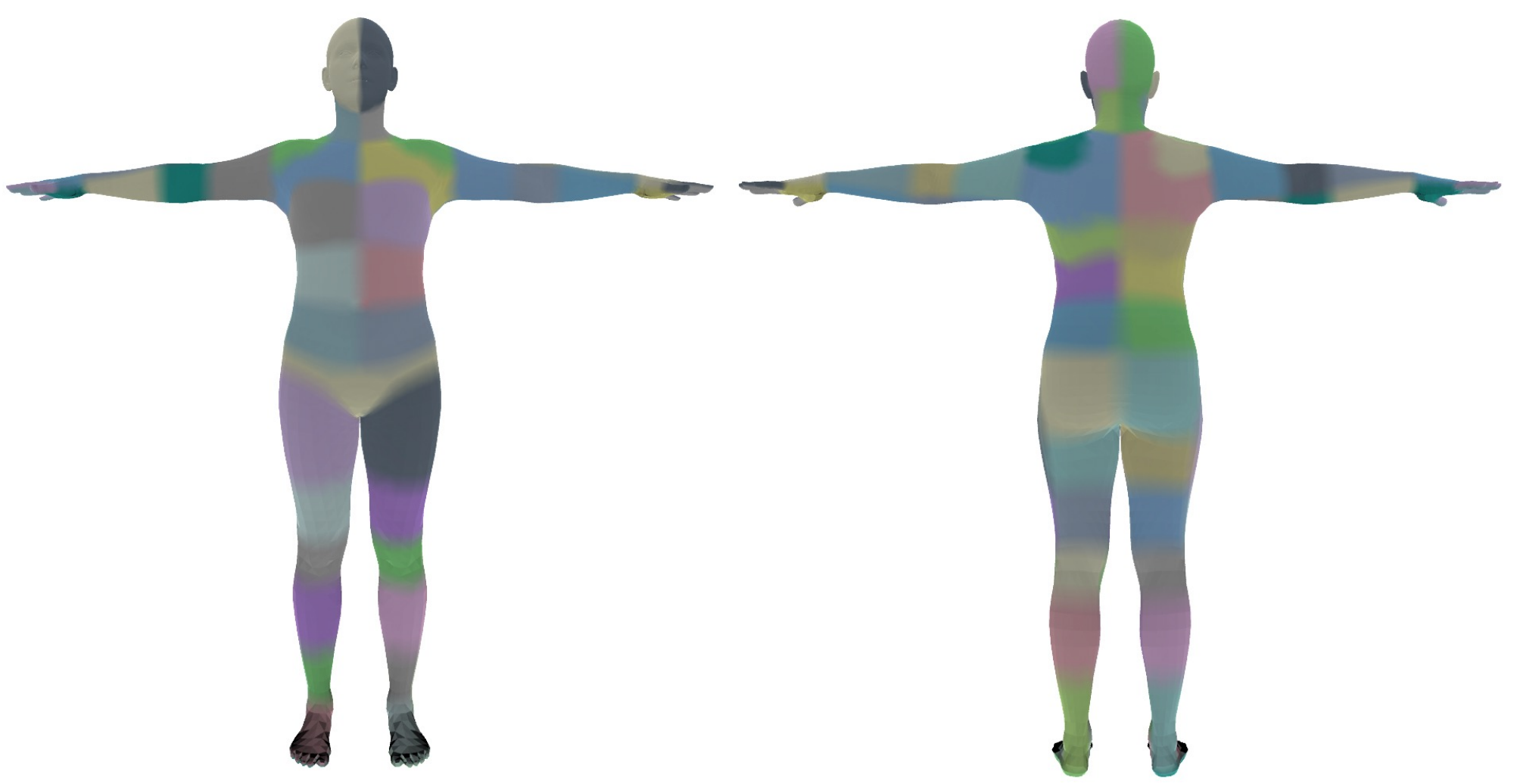}
    \caption{Color-coded 75 fine-grained regions used for PCC calculation}
    \label{fig:pcc-regions}
\end{figure*}
Figure~\ref{fig:pcc-regions} illustrates the 75 fine-grained regions used for PCC calculation, which are the same as those used in \cite{fieraru2020three}. We opted to compute PCC on the fine-grained regions rather than on the coarse ones since prior work uses the fine-grained regions \cite{muller2023generative} and since we want to measure contact correctness at a finer granularity (e.g. upper vs.~lower thigh vs.~knee). Since the regressors BEV and HMR2 use the SMPL mesh while the fine-grained regions are defined on the SMPL-X mesh, we use a matrix $M \in \mathbb{R}^{\text{num\_vertices\_smplx} \times \text{num\_vertices\_smpl}}$ to convert the SMPL meshes to SMPL-X in order to compute PCC.
\begin{table*}[t]
\centering 
\setlength{\tabcolsep}{6pt} 
\footnotesize
\begin{tabular}{@{}lrcrcr@{}}
\toprule 
 & \multicolumn{1}{c}{Hi4D} && \multicolumn{1}{c}{FlickrCI3D} && \multicolumn{1}{c}{CHI3D} \\
& $\text{PA-MPJPE}_{\downarrow}$ && $\text{PA-MPJPE}_{\downarrow}$ && $\text{PA-MPJPE}_{\downarrow}$ \\
\cmidrule{2-2} \cmidrule{4-4} \cmidrule{6-6}
\emph{Without contact supervision} \\
BEV~\cite{sun2022putting} & 76 && 71 && 51 \\
Heuristic & 65 && 31 && 48 \\
\methodname & 65 && 31 && 49 \\
\cmidrule{2-2} \cmidrule{4-4} \cmidrule{6-6}
\emph{With contact supervision} \\
BUDDI~\cite{muller2023generative} & 70 && 43 && 47 \\
\bottomrule
\end{tabular}
\caption{\textbf{Two-person Results.} Per-person PA-MPJPE (lower is better). For FlickrCI3D, PA-MPJPE is computed using the pseudo-ground-truth fits.}
\label{tab:perperson-pampjpe}
\vspace{-2mm}
\end{table*}
\subsection{Additional Quantitative Results}
\label{sec:supp-quant}
\subsubsection{Per-person PA-MPJPE}
Table~\ref{tab:perperson-pampjpe} shows the per-person PA-MPJPE for each of the datasets used in our two-person experiments.

\subsubsection{LMM Analysis}
\begin{table}[ht]
\centering 
\footnotesize
\setlength{\tabcolsep}{6.5pt}.
\begin{tabular}{@{}cccccc@{}}
\toprule 
 \multicolumn{1}{c}{LMM} & \multicolumn{1}{p{1.3cm}}{\centering Prompt \\ Style} & 
  \multicolumn{1}{p{0.8cm}}{\centering Ask for\\Left/right} & 
  \multicolumn{1}{p{0.8cm}}{\centering P1/P2 \\ Labeled?} & \multicolumn{1}{c}{$\text{Hi4D}_{\downarrow}$} & \multicolumn{1}{c}{$\text{Flickr}_{\downarrow}$} \\
 \cmidrule{1-6}
Heuristic & -- & -- & -- & $124$ & $67$ \\
\makecell*[l]{GT prob.\\random} & -- & -- & -- & $101$ & $68$ \\
\makecell*[l]{GT prob.\\random\\(no L/R)} & -- & -- & -- & $110$ & $61$ \\
GPT4-V & Direct & No & No & $83$ & $58$ \\
GPT4-V & Direct & Yes & No & $80$ & $62$ \\
GPT4-V & Direct & Yes & Yes & $82$ & $65$ \\
GPT4-V & Caption & No & No & $84$ & $60$ \\
GPT4-o & Direct & No & No & $84$ & $57$ \\ 
LLaVA & Direct & No & No & $86$ & $67$ \\
LLaVA & Caption & No & No & $89$ & $61$ \\
\bottomrule
\end{tabular}
\caption{\textbf{LMM Analysis}: 
We compare the default LMM and prompt (line 2) with several variants. We consider two other LMMs: GPT4-o (line 6), which refers to the \texttt{gpt-4o-08-06} version, and LLaVA v1.6 34B \citep{liu2024llavanext} (lines 7-8), and three other prompt variants: asking for left/right labels on limbs (line 3); asking for left/right labels on limbs and labeling the people in the image (line 4); and first generating a pose caption, removing mentions of left/right, and then converting the caption to constraints with a text-only LM (lines 6, 8). We take $N=5$ samples from each LMM. Metric is PA-MPJPE on Hi4D/FlickrCI3D validation sets.} 
\label{tab:lmm-analysis}
\end{table}
In Table~\ref{tab:lmm-analysis}, we show that \methodname works with other LMMs--performance is comparable with GPT-4o and worse with LLaVA, which is in line with the general capabilities of these LMMs. For details on how we use LLaVA and other LLaVA results, see \S~\ref{sec:supp-llava}. We also compare the default prompt with three other prompt types: (1) asking for left/right labels on each limb, (2) labeling each person on the image and asking for left/right labels on each limb, and (3) generating a pose-focused caption, removing mentions of ``left''/``right'' from it by feeding it to a language model, and converting the resulting caption to a set of constraints with another call to a language model. When converting the constraints from Prompt 3 into programs, we find that programs can become prohibitively long, so we discard loss functions that are longer than 10000 characters. Prompt 3 resembles the classify-then-constrain pipeline of \cite{humanobjectlanguage}. When using GPT4-V, the default prompt outperforms Prompt 3 on both datasets and Prompts 1/2 on at least one (see \S~\ref{sec:limitations} for further discussion). In addition to the heuristic, Table~\ref{tab:lmm-analysis} includes another baseline which for a given image decides to include each possible constraint pair $p$ with probability $C_p$/$|D|$, where $C_p$ is the number of occurrences of $p$ in the dataset and $|D|$ is the number of images in the dataset. We include two variants of the baseline, one where left/right labels are retained and another where they are removed and both chiralities are considered in the loss function (as in ProsePose). The PA-MPJPE of ProsePose with GPT4-V and the default prompt is better than the PA-MPJPE of both baselines. 

\subsubsection{Running Time}
We compare BUDDI and BUDDI+ProsePose on Hi4D val. The average time per example is 64 sec. for BUDDI vs. 89 sec. for BUDDI+ProsePose. The time to sample 20 programs from GPT-4o, averaged over 30 examples, is 16 sec.

\subsubsection{Variance across Camera Angles}
We quantify the impact of the camera angle on LMM predictions by running GPT-4o (the \texttt{gpt-4o-08-06} version) with each camera angle. The F1 ranges from 0.31 to 0.42, but the only cameras with F1 less than 0.37 are the front and back cameras. Some body parts are often occluded from these angles, which may explain the lower scores.

\subsection{LLaVA Results}
\label{sec:supp-llava}
In this section, we provide test set results when LLaVA-NeXT 34B (i.e. LLaVA v1.6) \cite{liu2023llava} is used as the LMM. 
We use the caption-to-table prompting approach described in the prompt ablation study (\S~\ref{sec:results-two}). That is, we generate a caption from the LMM, and we feed the caption alone to GPT4 in order to convert it into a table of constraints.
For the two-person case, the prompts are given in \S\ref{sec:supp-ablation-prompts}. We use a temperature of 0.3 and top-p of 0.7 when sampling from LLaVA.

For the one-person case, we use the following prompt for LLaVA:
\begin{tcolorbox}
Describe the person's pose.
\end{tcolorbox}
We use the same prompt as above to rewrite the caption. We then use the following prompt to create the formatted table:
\begin{tcolorbox}
You are a helpful assistant. You will follow ALL rules and directions entirely and precisely.

Given a description of a yoga pose, create a Markdown table with the columns "Body Part 1" and "Body Part 2", listing the body parts of the person that are guaranteed to be in contact with each other, from the following list. ALL body parts that you list must be from this list.
Body parts: "head", "back", "shoulder", "arm", "hand", "leg", "foot", "stomach", "butt", "ground"
Note that "back" includes the entire area of the back.

Include all contact points that are directly implied by the description, not just those that are explicitly mentioned.
If there are no contact points between these body parts that the description implicitly or explicitly implies, your table should contain only the column names and no other rows.

First, write your reasoning. Then write the Markdown table.
\end{tcolorbox}
\begin{table*}[t]
\centering 
\setlength{\tabcolsep}{6pt} 
\footnotesize
\begin{tabular}{@{}lrrcrrrcrrrcrrr@{}}
\toprule 
 & \multicolumn{2}{c}{Hi4D} && \multicolumn{3}{c}{FlickrCI3D} && \multicolumn{3}{c}{CHI3D} && \multicolumn{3}{c}{MOYO} \\
& $\text{Err}_{\downarrow}$ & $\text{F1}_{\uparrow}$ && $\text{Err}_{\downarrow}$ & $\text{PCC}_{\uparrow}$ & $\text{F1}_{\uparrow}$ && $\text{Err}_{\downarrow}$ & $\text{PCC}_{\uparrow}$ & $\text{F1}_{\uparrow}$ && $\text{Err}_{\downarrow}$ & $\text{PCC}_{\uparrow}$ & $\text{F1}_{\uparrow}$\\
\cmidrule{2-3} \cmidrule{5-7} \cmidrule{9-11} \cmidrule{13-15}
Heuristic & 116 & -- && 67 & 77.8 & -- && 105 & 74.1 & -- && -- & -- & --\\
HMR2+opt & -- & -- && -- & -- & -- && -- & -- & -- && 81 & 85.2 & -- \\
GPT4-V & 93 & 24 && 58 & 79.9 & 13 && 100 & 75.8 & 23 && 82 & 87.8 & 25 \\
LLaVA+GPT4 & 95 & 22 && 60 & 79.7 & 7 && 101 & 75.2 & 13 && 82 & 85.2 & 14 \\
\bottomrule
\end{tabular}
\caption{\textbf{LLaVA Results.} Err denotes Joint PA-MPJPE for the two-person datasets (Hi4D, FlickrCI3D, CHI3D) and PA-MPJPE for MOYO. Lower is better for Err, and higher is better for Avg.~PCC. Note that the GPT4-V results use 20 samples from the LMM, while the LLaVA results use 5 samples from the LMM.} 
\label{tab:llava}
\vspace{-2mm}
\end{table*}
We set $N=5$ for these experiments.
Since we change $N$, we also need to select appropriate thresholds $f$ and $t$. As in the experiments with GPT4-V, we set $t = N$ for all datasets except CHI3D. For CHI3D, we find on the validation set that $t=2$ works better than $t=1$, so we set $t=2$. As in the experiments with GPT4-V, we set $f=1$ for the 2-person datasets, and we set $f=3$ for MOYO, to approximate the ratio $f/N$ used in the GPT4-V experiments. Finally, as stated in Appendix~\ref{sec:supp-quant}, we discard loss functions that are longer than 10000 characters.

Table~\ref{tab:llava} shows the results. On the 2-person datasets, the LLaVA+GPT4 approach performs better than the contact heuristic but not as well as GPT4-V. This is in line with holistic multimodal evaluations that indicate that GPT4-V performs better than LLaVA \cite{wildvision}. On the 1-person yoga dataset, the performance of LLaVA+GPT4 is comparable with that of the baseline (HMR2+opt). The reason that LLaVA performs worse than GPT4-V in this setting may be that LLaVA does not have enough training data on yoga to provide useful constraints.

\subsection{Failure cases}
\begin{figure*}
    \centering
    \includegraphics[width=\linewidth]{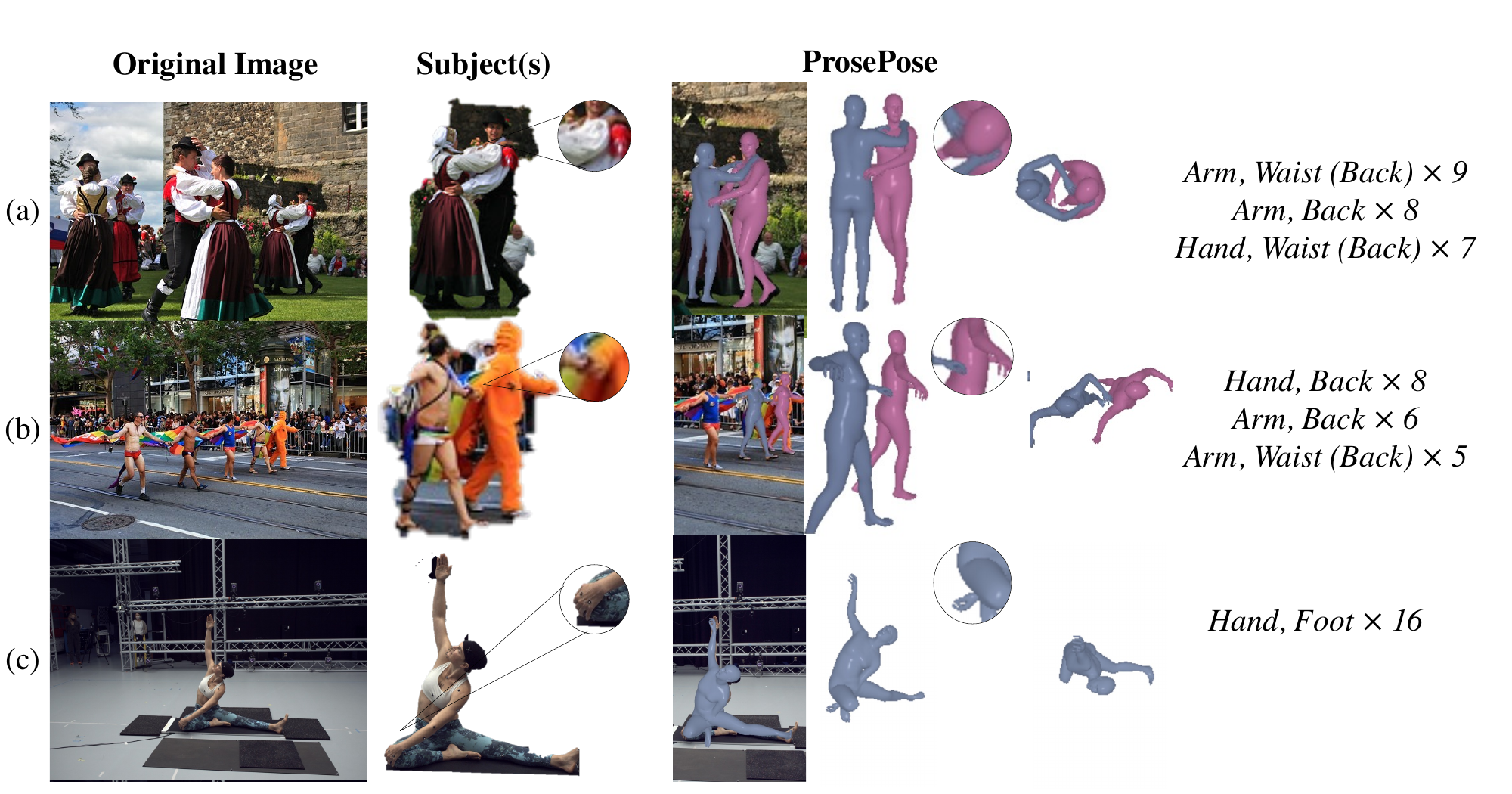}
    \vspace{-0.5cm}
    \caption{\textbf{Failure cases} We show examples in which \methodname fails to output a semantically correct pose. The constraints shown are the top 3 constraints (or the total number of constraints, whichever is smaller) that meet the threshold $f$ along with their counts ($f=1$ for two-person experiments and $f=10$ for the one-person experiment).}
    \label{fig:failures}
\end{figure*}

Figure~\ref{fig:failures} shows examples of two types of ProsePose failures: (1) incorrect chirality (example a) and (2) hallucination (examples b and c). In example (a), the top constraints are correct but without the chirality specified. The optimization then brings both hands of one person to roughly the same point on the other person's waist, rather than positioning one hand on each hip. Similarly, both hands of the other person are positioned on the same shoulder of the first person. Examples (b) and (c) both show cases of hallucination. In example (b), the hand is predicted to touch the back rather than the hand. In example (c), the hand is predicted to touch the foot rather than the leg. Interestingly, in the yoga example, GPT4-V correctly predicts the name of the yoga pose in all 20 samples (``Parivrtta Janu Sirsasana''). However, it outputs a constraint between a hand and a foot, which is true in the standard form of this pose but not in the displayed form of the pose. Consequently, the optimization brings the left hand closer to the right foot than to the right knee.

\begin{figure*}
    \centering
    \includegraphics[width=0.9\linewidth]{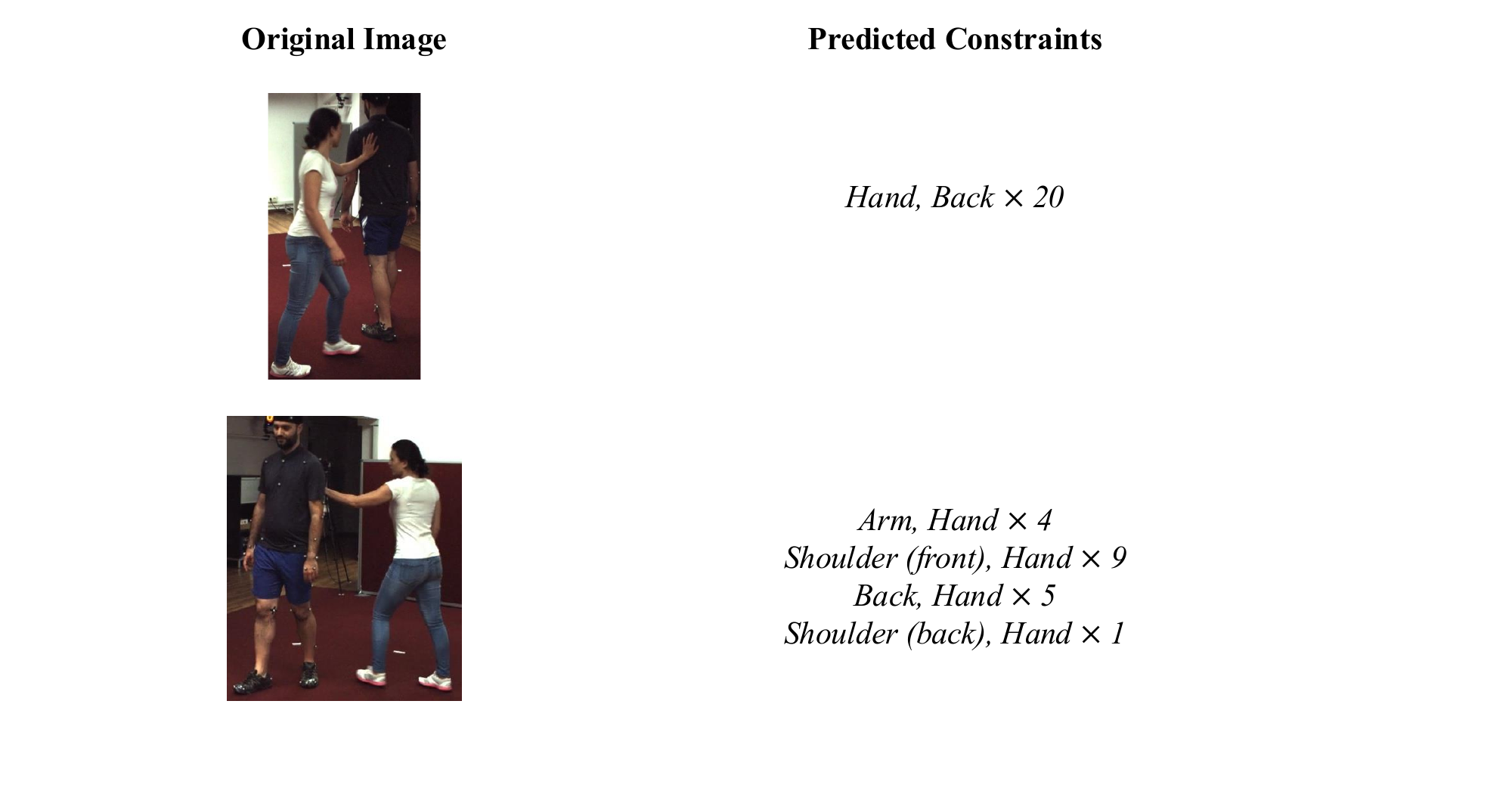}
    \vspace{-0.5cm}
    \caption{\textbf{Variation due to camera angle} Example from CHI3D in which GPT4-V outputs substantially different constraint sets for different views of the same pose.}
    \label{fig:failures-camera}
\end{figure*}

Figure~\ref{fig:failures-camera} shows an example in which the camera view affects GPT4-V's predictions substantially.

\begin{figure*}
    \centering
    \includegraphics[width=0.9\linewidth]{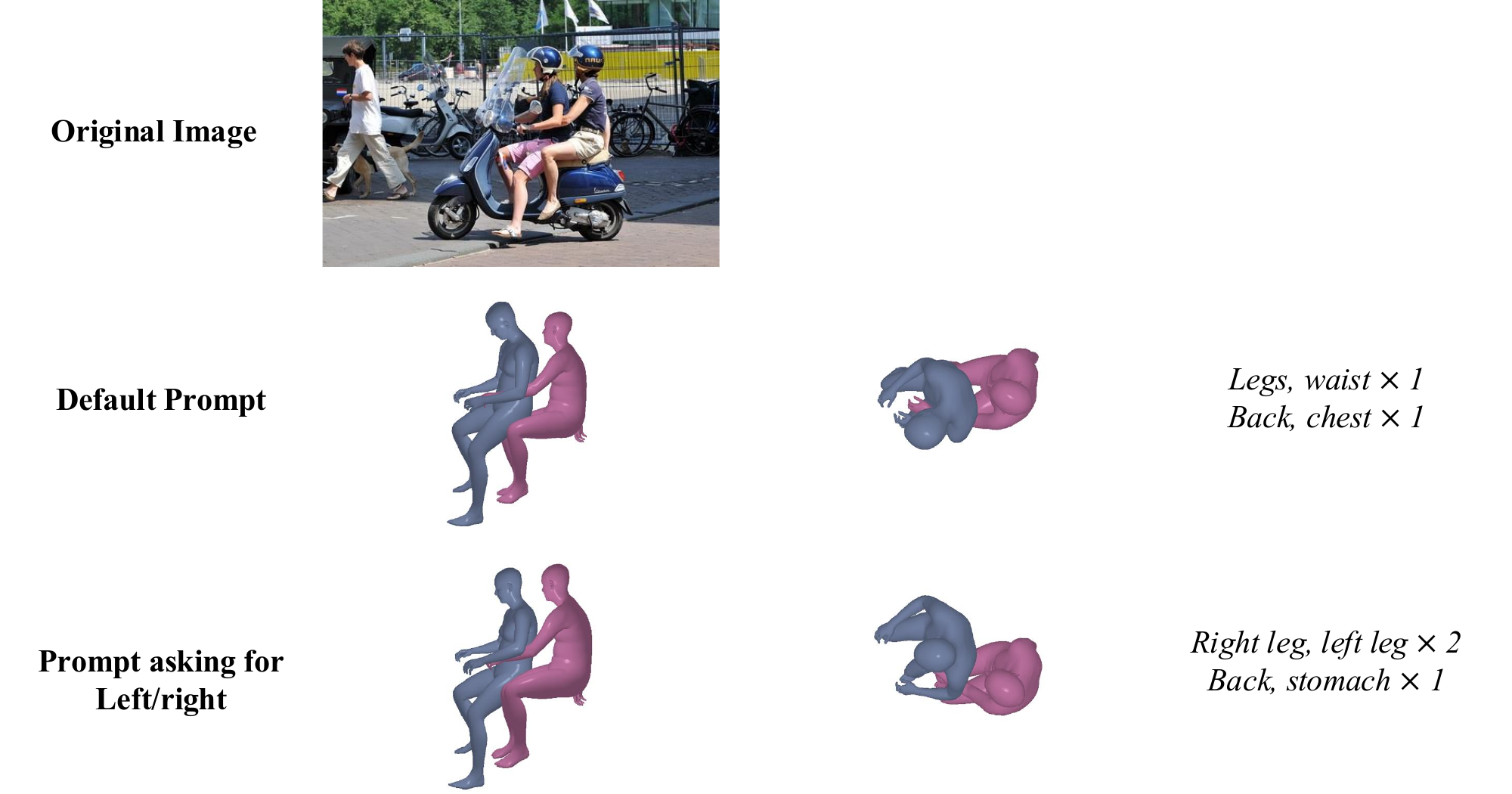}
    \vspace{-0.3cm}
    \caption{\textbf{Failure of left/right prompt} In this example, GPT4-V responds with an incorrect chirality when asked to specify the chirality of the limbs in the constraints. The last column shows the unordered valid region pairs occurring in the 5 samples from GPT4-V along with the number of samples in which the pair occurs.}
    \label{fig:failures2}
\end{figure*}

\begin{figure*}
    \centering
    \includegraphics[width=0.9\linewidth]{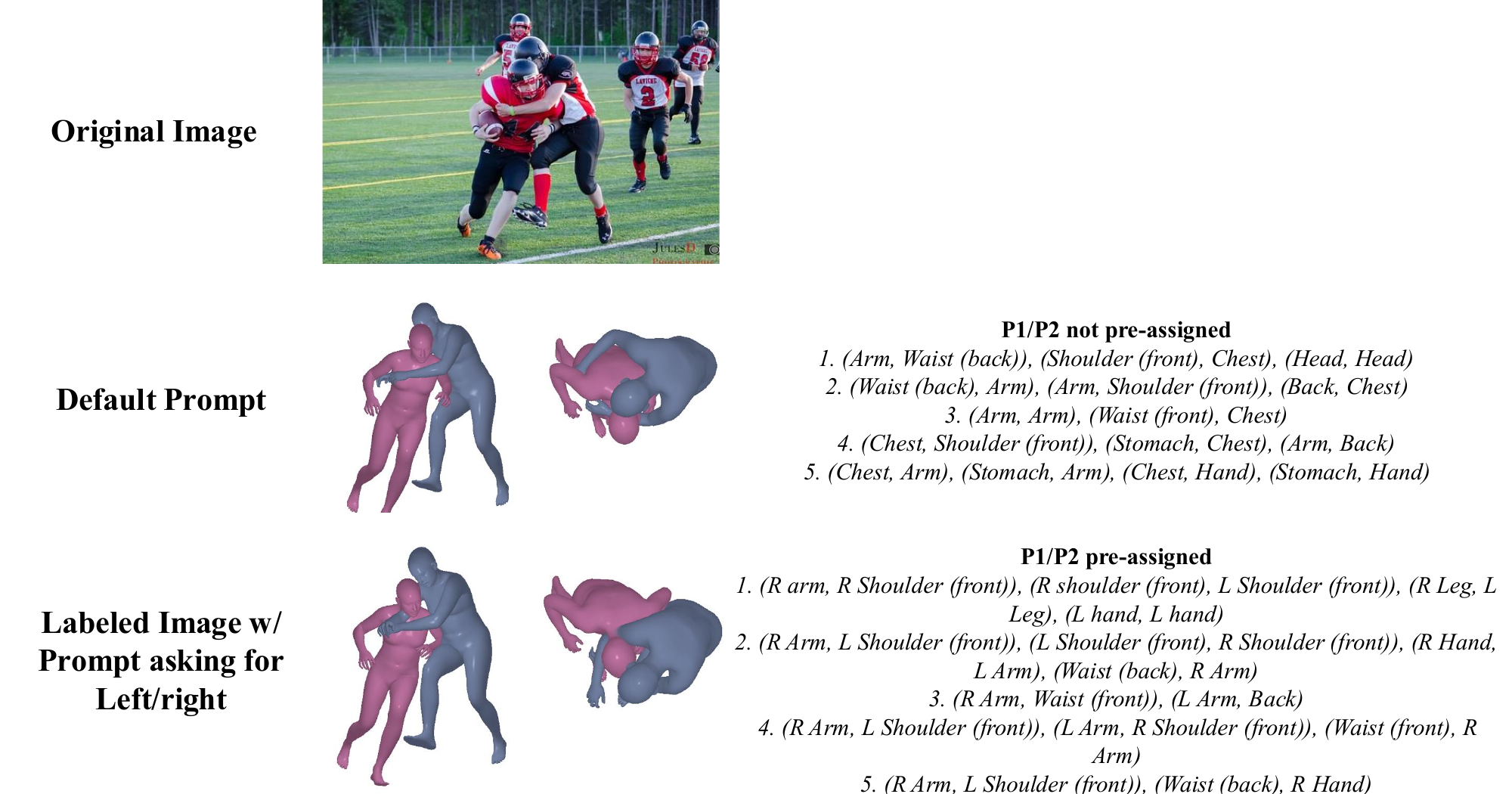}
    \caption{\textbf{Failure of labeled image with left/right prompt} In this example, the LMM responds with incorrect constraints when given an image in which persons 1 and 2 are labeled and asked to specify the chirality of the limbs in the constraints. The last column shows the constraint sets in the 5 LMM samples. For the default prompt, since persons 1 and 2 are not labeled in the image, the minimum loss from the two labelings is used, while for the other prompt, only the loss corresponding to the given labeling is considered.}
    \label{fig:failures3}
\end{figure*}

Given that the chirality is an important issue, it is natural to consider a prompt that asks the LMM to specify a chirality for each limb. Table~\ref{tab:lmm-analysis} shows that such a prompt does not outperform our default prompt. Figure~\ref{fig:failures2} shows an example of a failure of the prompt that requests chirality. With the prompt that asks for chirality, GPT4-V incorrectly predicts that the right leg of one person is touching the left leg of the other person. With the default prompt, GPT4-V predicts in one constraint set that one person's legs and chest are touching the other's waist and back, respectively (and empty constraint sets otherwise). In the prompt ablation study, we also consider a prompt in which the image is labeled with Person 1 and Person 2 and the prompt asks for left/right labels, and this prompt also does not outperform the default one. Figure~\ref{fig:failures3} shows an example in which the alternative prompt leads to incorrect predictions from GPT4-V.

\label{sec:supp-failures}

\subsection{Additional Qualitative Results}
\label{sec:supp-qual}
\begin{figure*}
    \centering
    \begin{sideways}
        \includegraphics[scale=0.28]{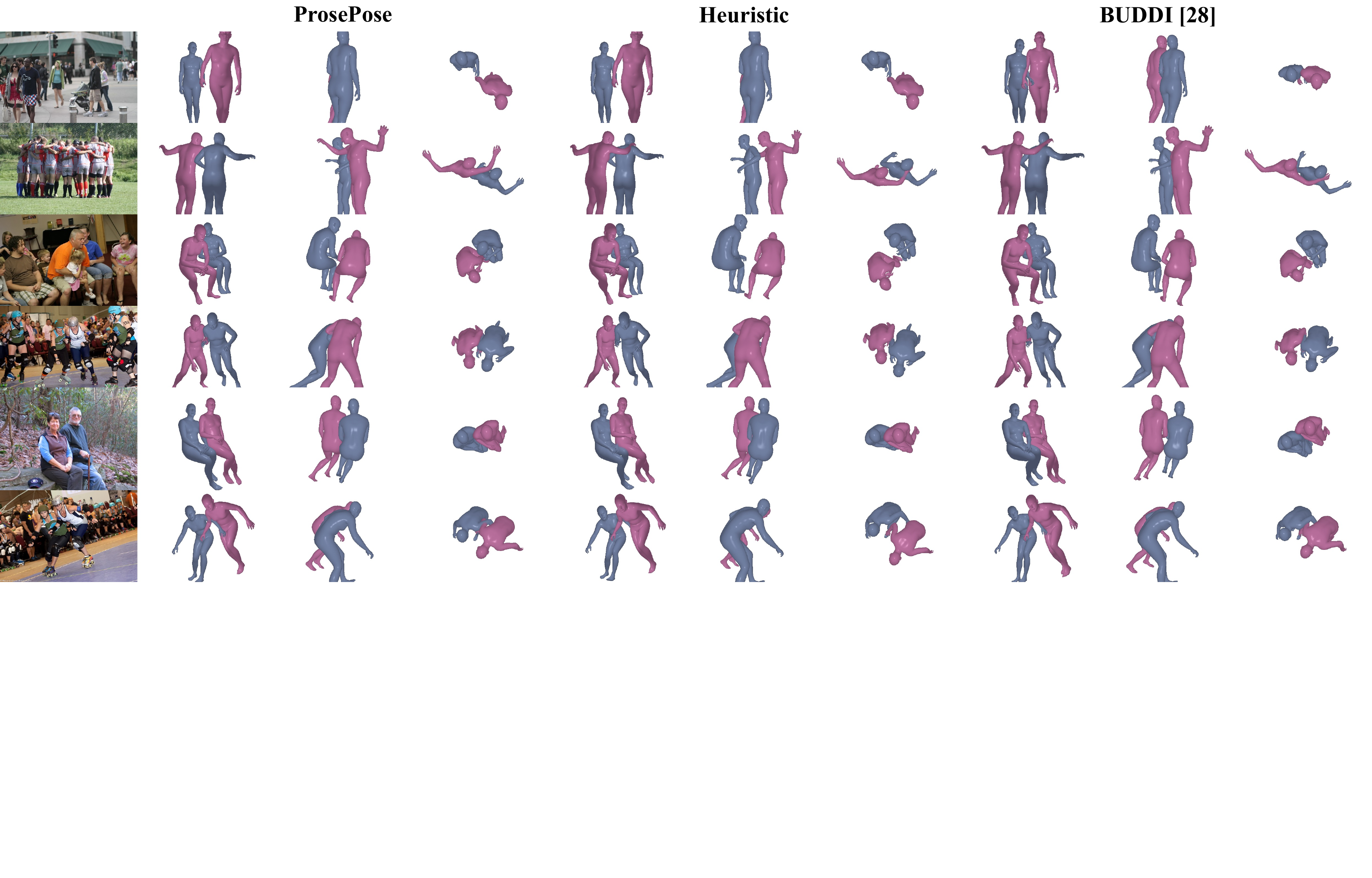}
    \end{sideways}
    \caption{Non-curated examples from the FlickrCI3D test set. They are randomly selected from the examples for which there is at least one non-empty constraint set.}
    \label{fig:supp-two-person-part1}
\end{figure*}

\begin{figure*}
    \centering
    \begin{sideways}
        \includegraphics[scale=0.28]{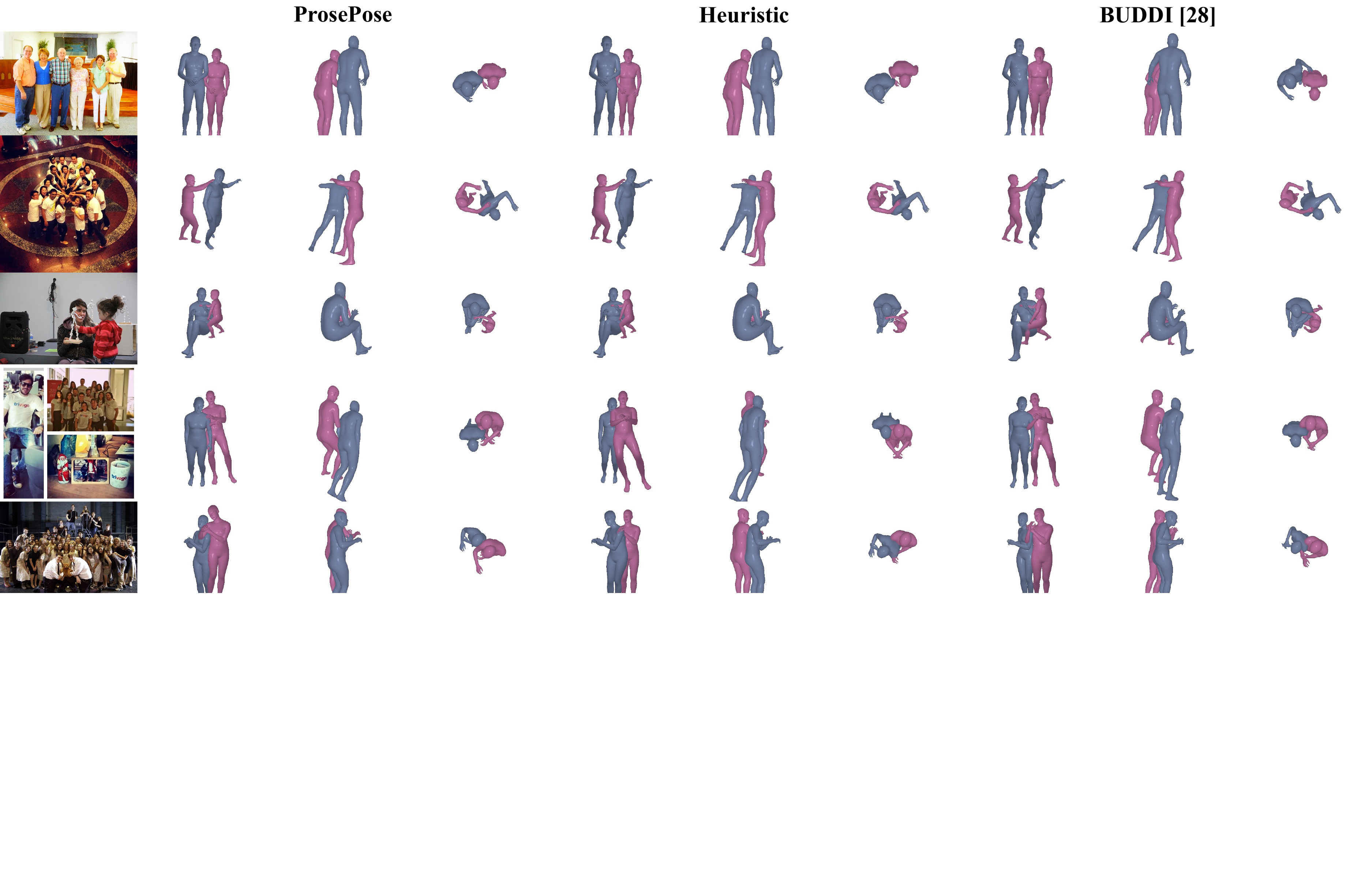}
    \end{sideways}
    \caption{Non-curated examples from the FlickrCI3D test set. They are randomly selected from the examples for which there is at least one non-empty constraint set.}
    \label{fig:supp-two-person-part2}
\end{figure*}

\begin{figure*}
    \centering
    \begin{sideways}
        \includegraphics[scale=0.28]{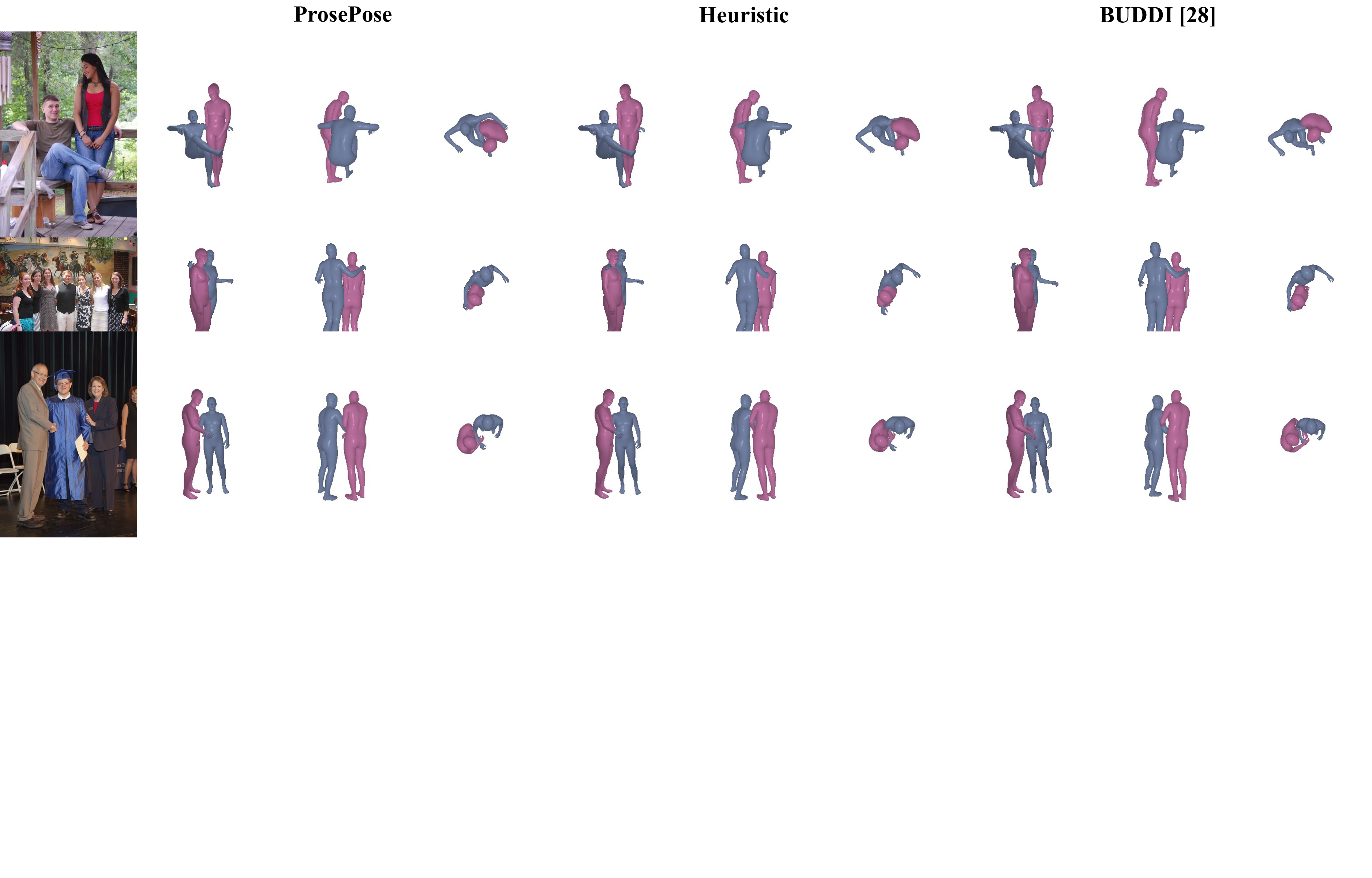}
    \end{sideways}
    \caption{Non-curated examples from the FlickrCI3D test set. They are randomly selected from the examples for which there is at least one non-empty constraint set.}
    \label{fig:supp-two-person-part3}
\end{figure*}

\begin{figure*}
    \centering
    \begin{sideways}
        \includegraphics[scale=0.28]{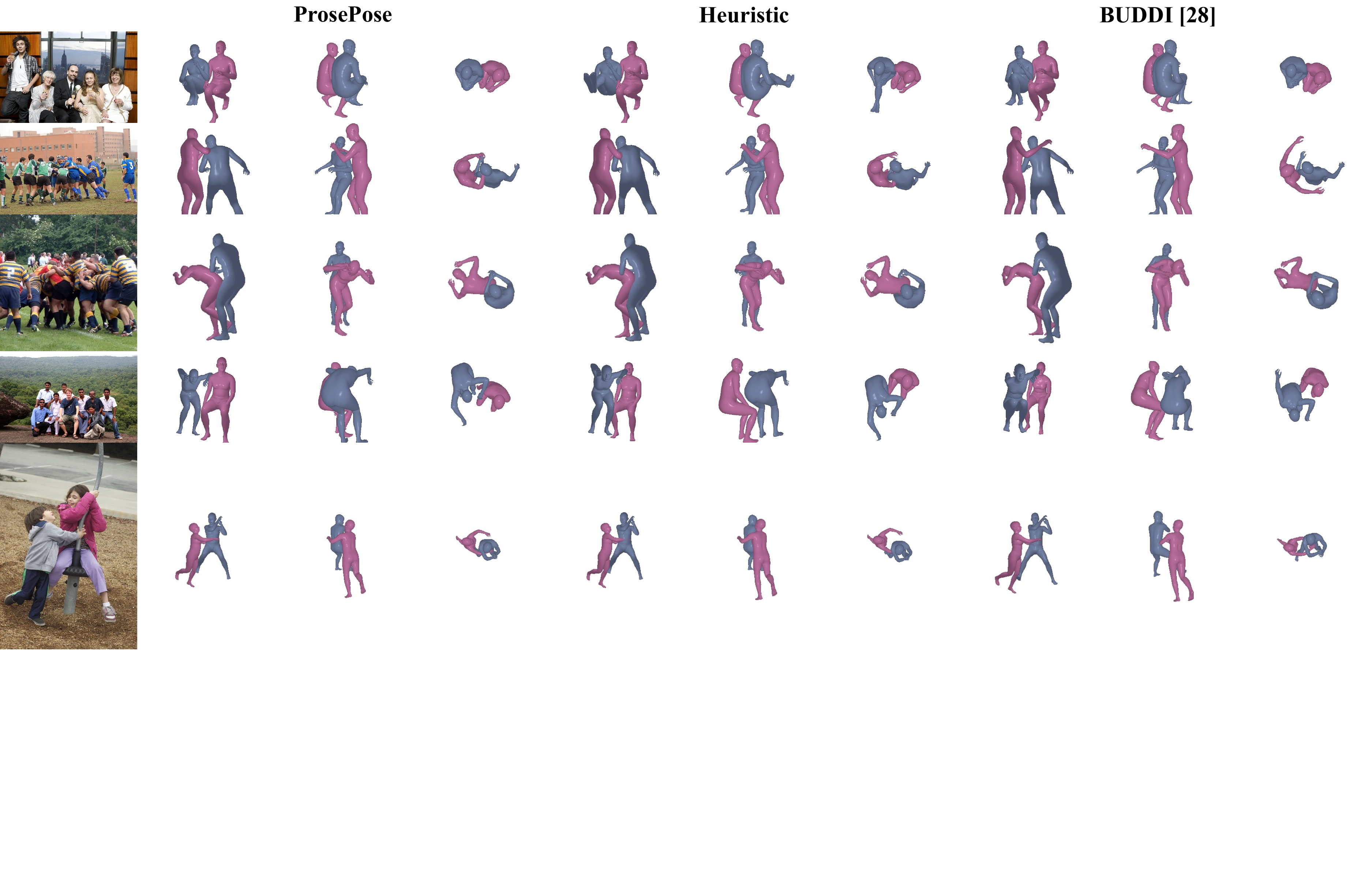}
    \end{sideways}
    \caption{Non-curated examples from the FlickrCI3D test set. They are randomly selected from the examples for which there is at least one non-empty constraint set.}
    \label{fig:supp-two-person-part4}
\end{figure*}

\begin{figure*}
    \centering
    \begin{sideways}
        \includegraphics[scale=0.28]{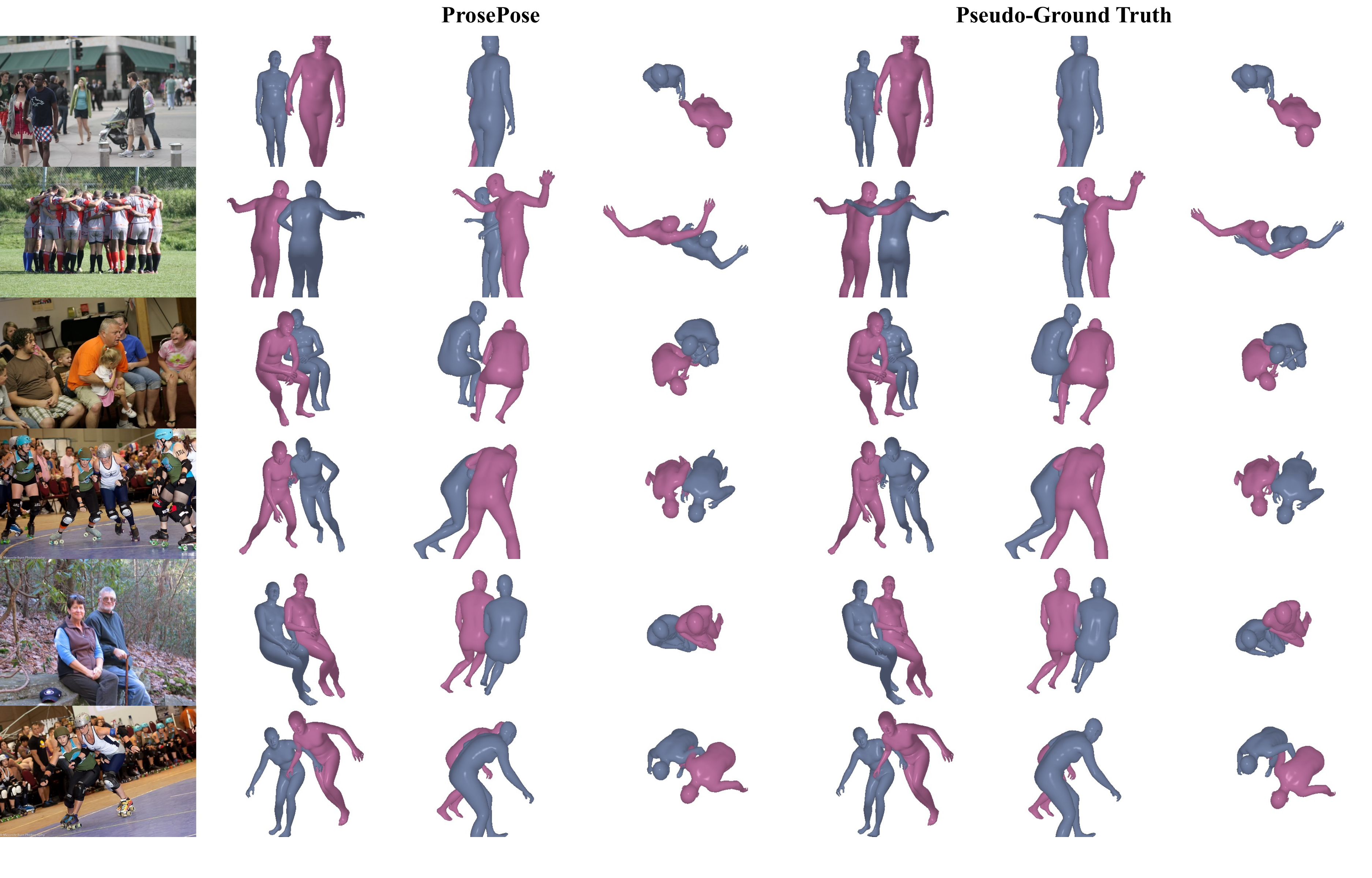}
    \end{sideways}
    \caption{Non-curated examples from the FlickrCI3D test set, comparing \methodname with the pseudo-ground truth fits. They are randomly selected from the examples for which there is at least one non-empty constraint set.}
    \label{fig:supp-pgt-part1}
\end{figure*}

\begin{figure*}
    \centering
    \begin{sideways}
        \includegraphics[scale=0.28]{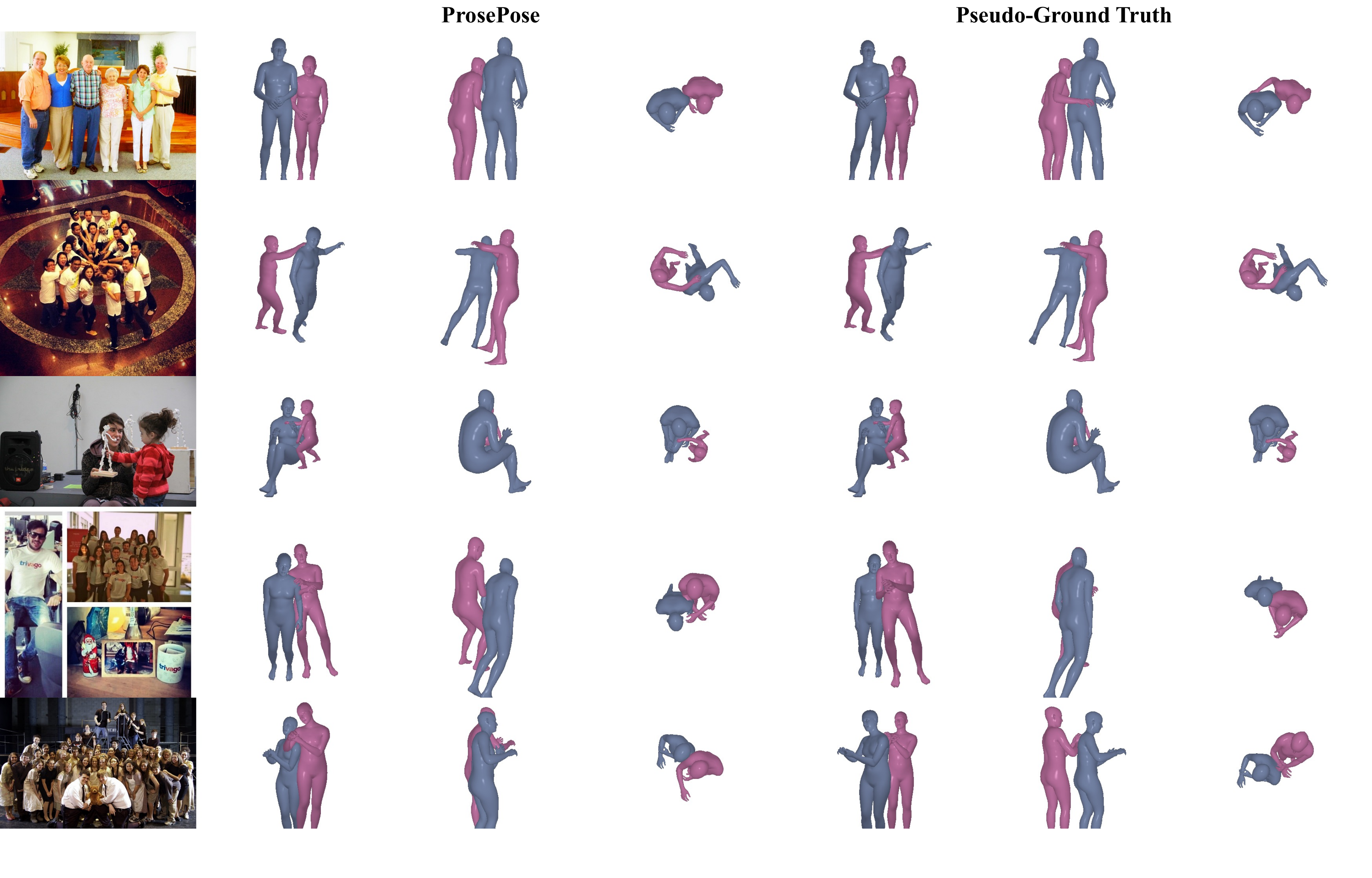}
    \end{sideways}
    \caption{Non-curated examples from the FlickrCI3D test set, comparing \methodname with the pseudo-ground truth fits. They are randomly selected from the examples for which there is at least one non-empty constraint set.}
    \label{fig:supp-pgt-part2}
\end{figure*}

\begin{figure*}
    \centering
    \begin{sideways}
        \includegraphics[scale=0.28]{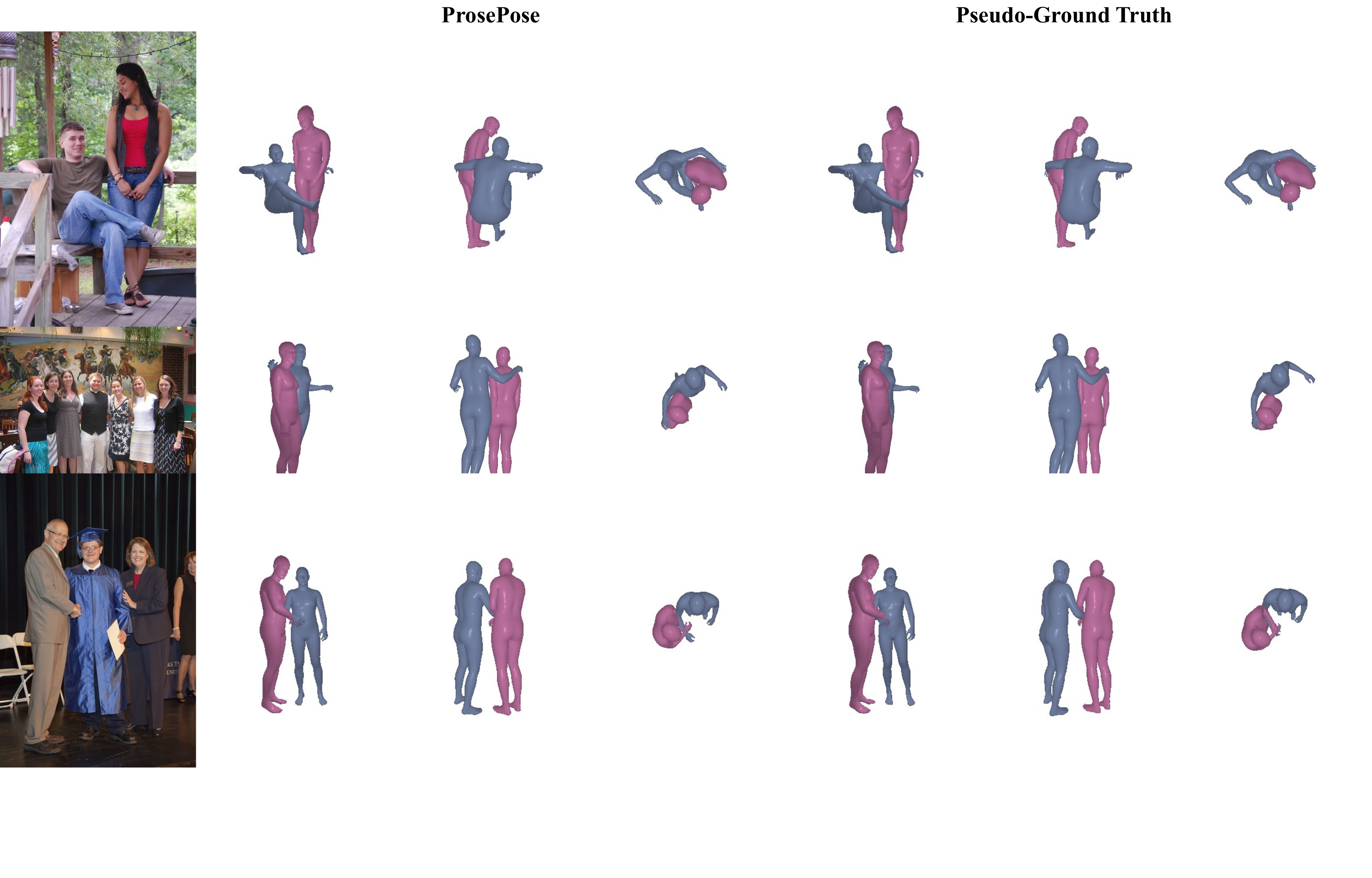}
    \end{sideways}
    \caption{Non-curated examples from the FlickrCI3D test set, comparing \methodname with the pseudo-ground truth fits. They are randomly selected from the examples for which there is at least one non-empty constraint set.}
    \label{fig:supp-pgt-part3}
\end{figure*}

\begin{figure*}
    \centering
    \begin{sideways}
        \includegraphics[scale=0.28]{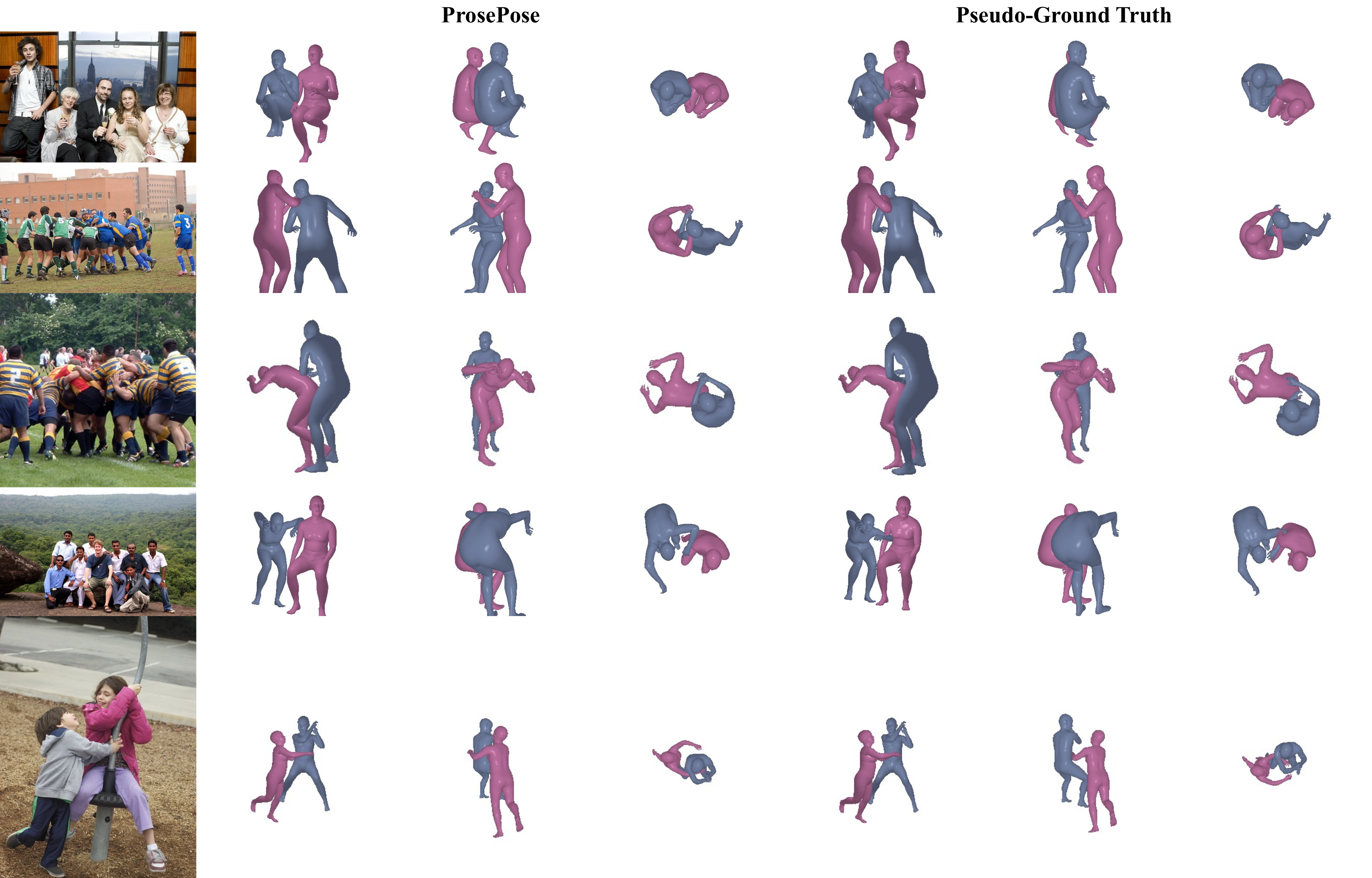}
    \end{sideways}
    \caption{Non-curated examples from the FlickrCI3D test set, comparing \methodname with the pseudo-ground truth fits. They are randomly selected from the examples for which there is at least one non-empty constraint set.}
    \label{fig:supp-pgt-part4}
\end{figure*}

\begin{figure*}
    \centering
    \begin{sideways}
        \includegraphics[scale=0.28]{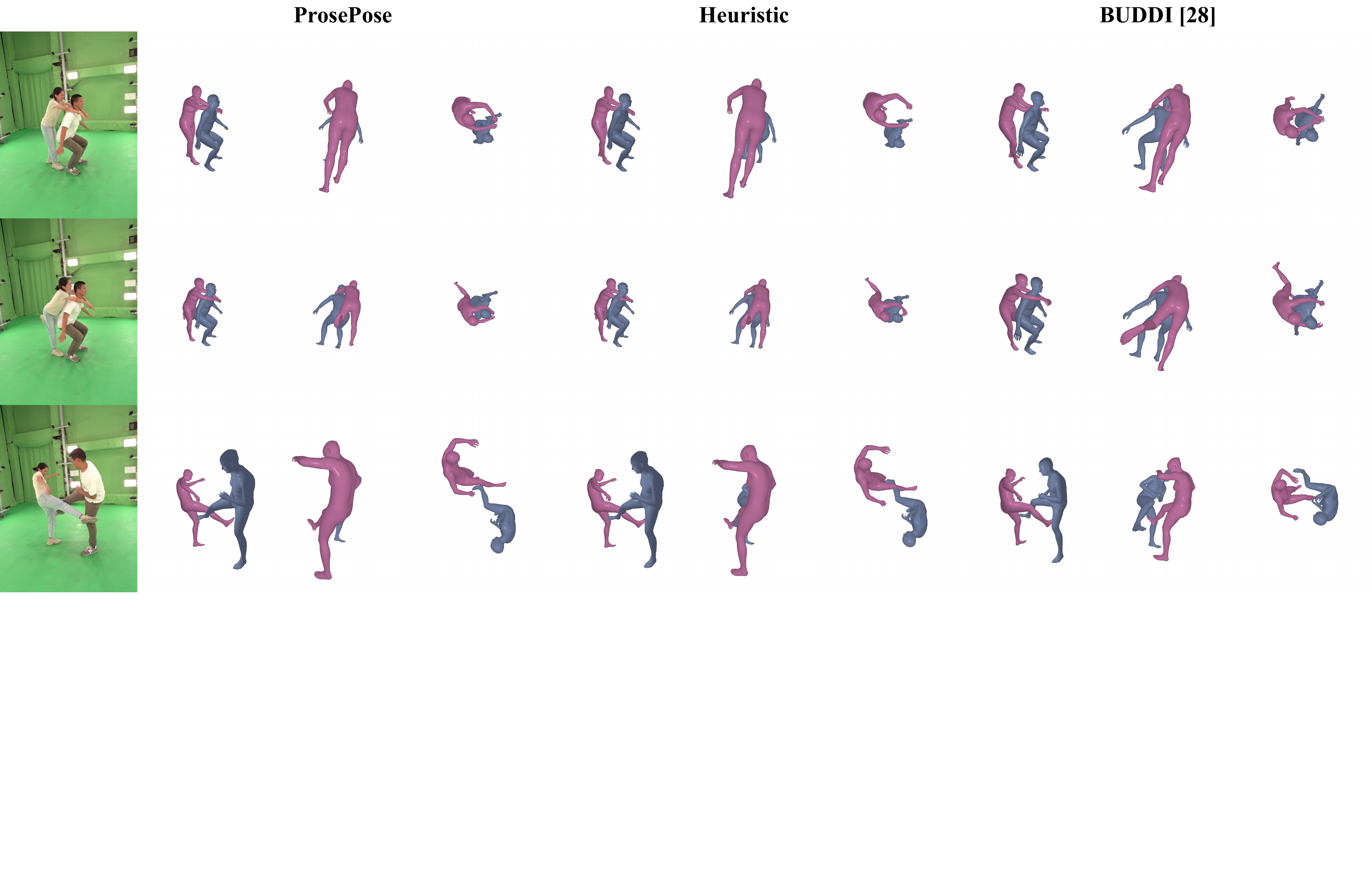}
    \end{sideways}
    \caption{Non-curated examples from the Hi4D test set. They are randomly selected from the examples for which there is at least one non-empty constraint set.}
    \label{fig:supp-hi4d-part1}
\end{figure*}

\begin{figure*}
    \centering
    \begin{sideways}
        \includegraphics[scale=0.28]{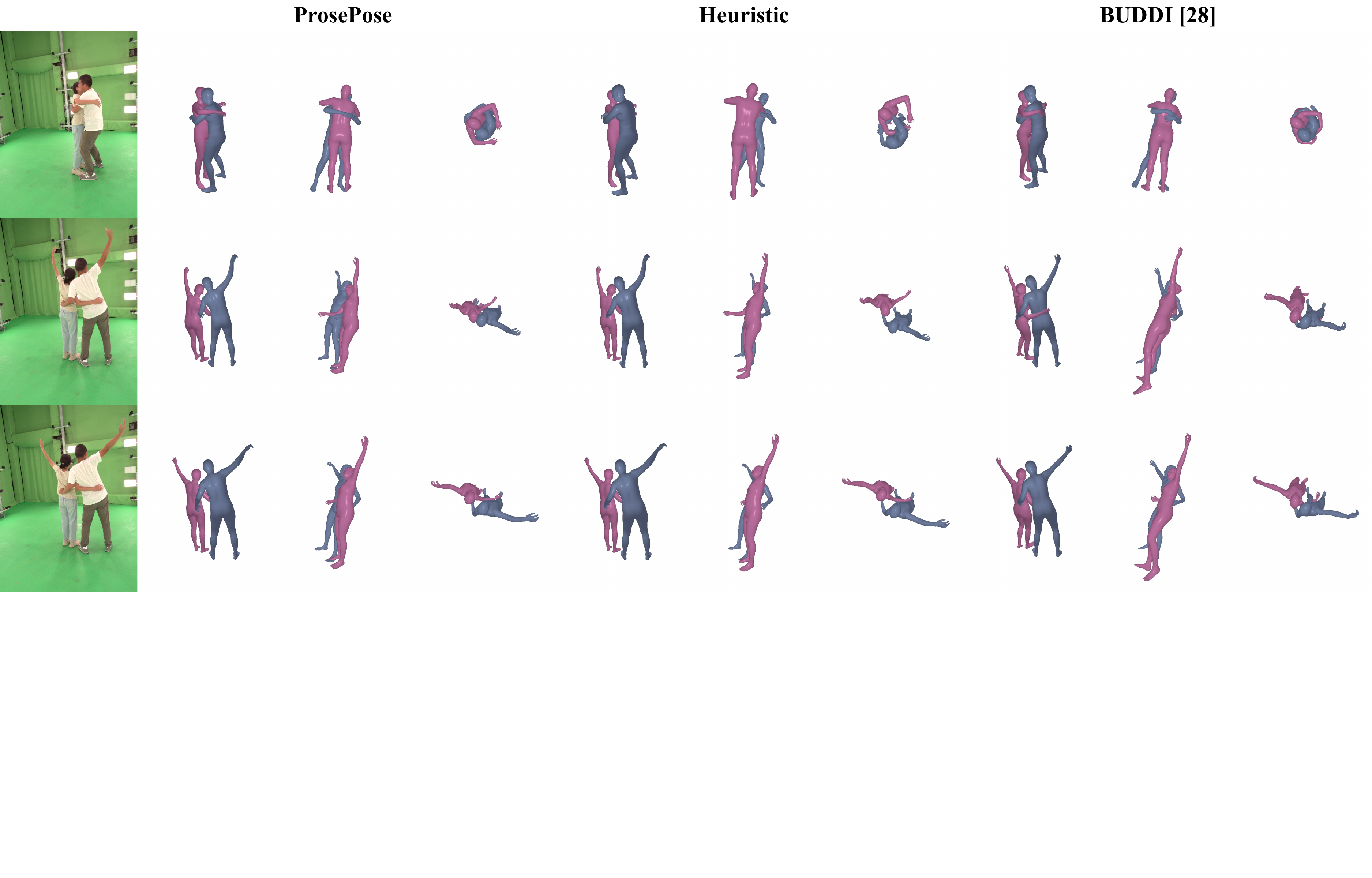}
    \end{sideways}
    \caption{Non-curated examples from the Hi4D test set. They are randomly selected from the examples for which there is at least one non-empty constraint set.}
    \label{fig:supp-hi4d-part2}
\end{figure*}

\begin{figure*}
    \centering
    \begin{sideways}
        \includegraphics[scale=0.28]{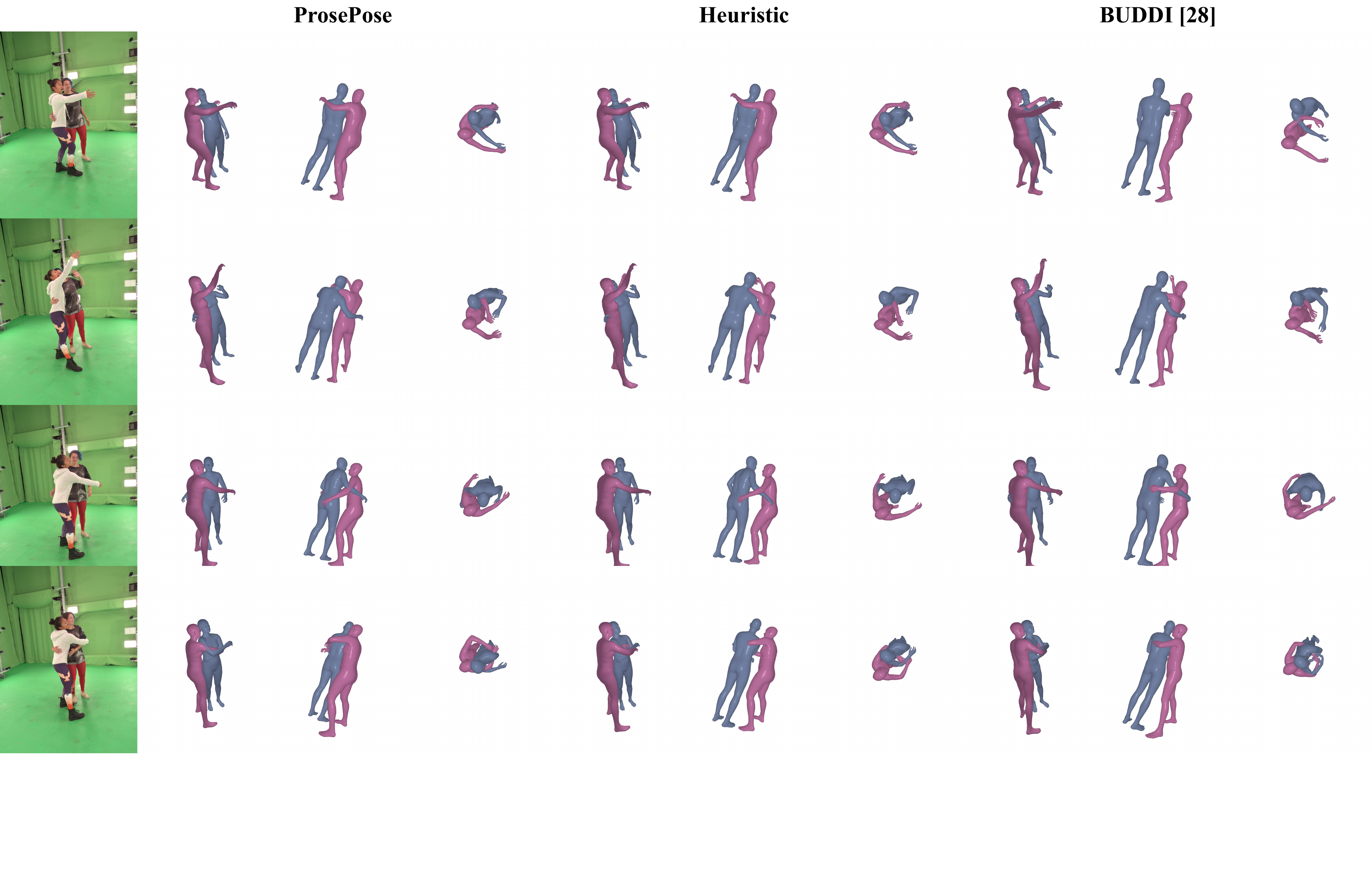}
    \end{sideways}
    \caption{Non-curated examples from the Hi4D test set. They are randomly selected from the examples for which there is at least one non-empty constraint set.}
    \label{fig:supp-hi4d-part3}
\end{figure*}

\begin{figure*}
    \centering
    \begin{sideways}
        \includegraphics[scale=0.28]{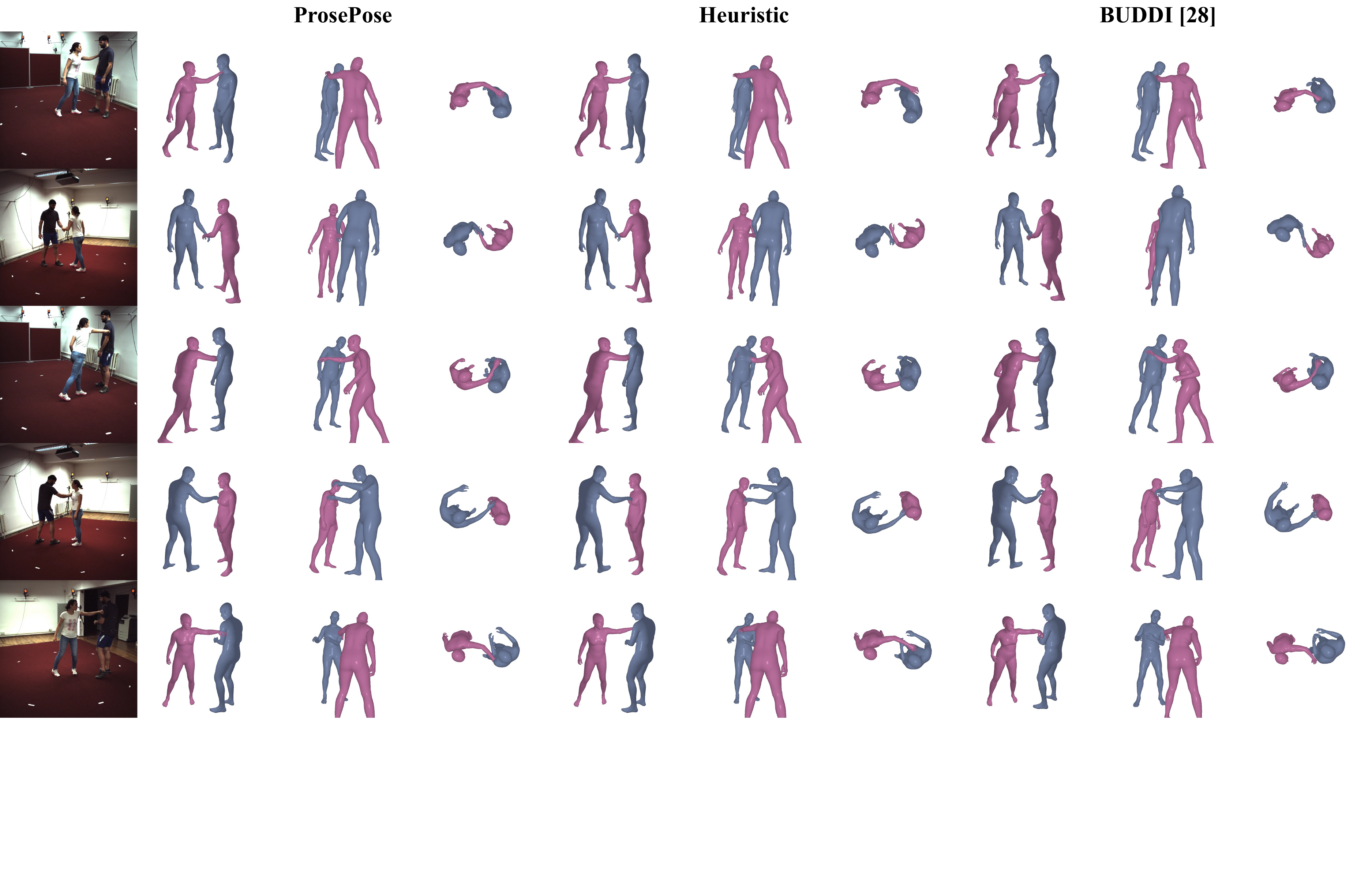}
    \end{sideways}
    \caption{Non-curated examples from the CHI3D validation set (which we use as the test set). They are randomly selected from the examples for which there are at least nineteen non-empty constraint sets (since we set $t=2$ for CHI3D).}
    \label{fig:supp-chi3d-part1}
\end{figure*}

\begin{figure*}
    \centering
    \begin{sideways}
        \includegraphics[scale=0.28]{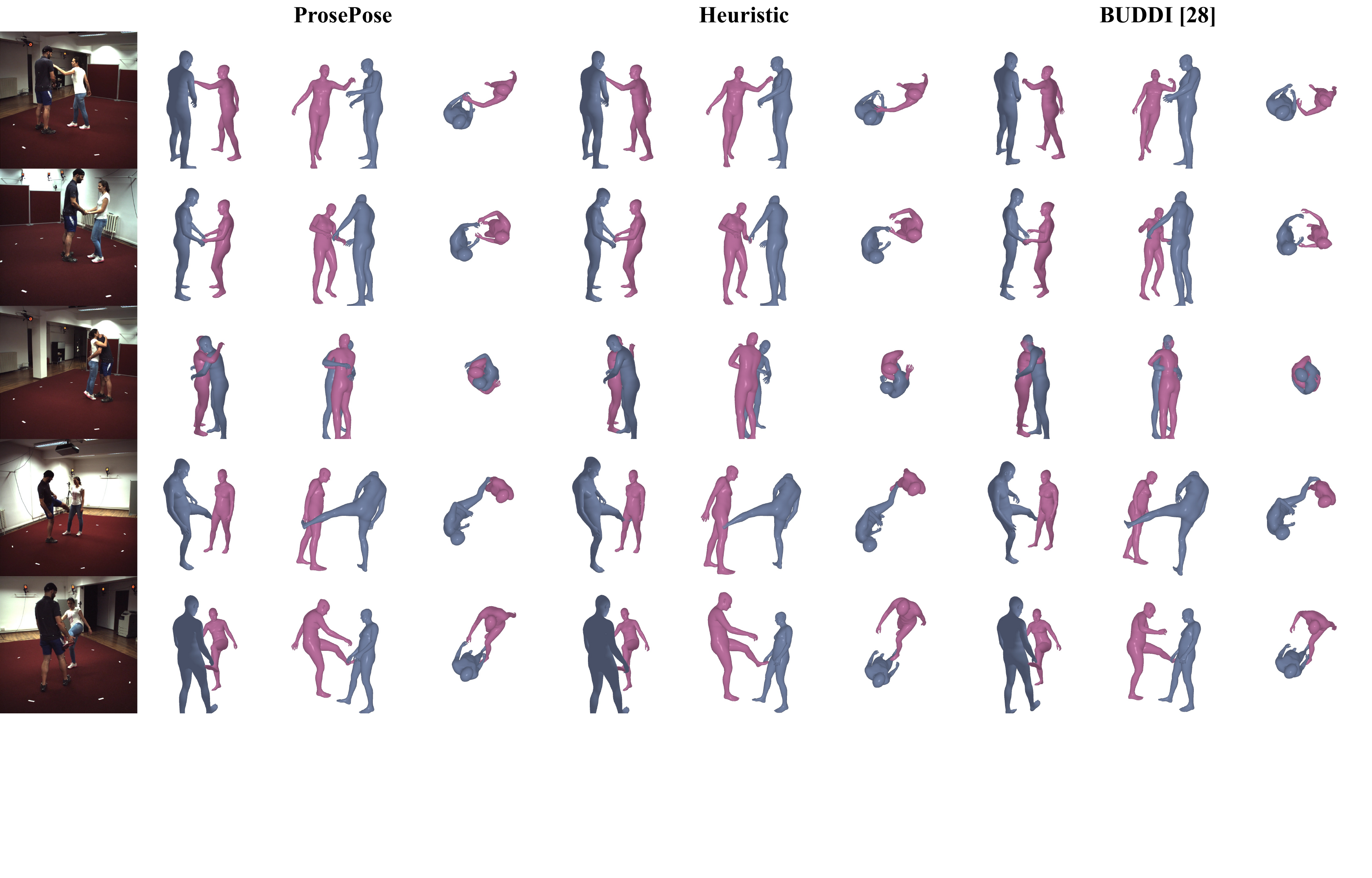}
    \end{sideways}
    \caption{Non-curated examples from the CHI3D validation set (which we use as the test set). They are randomly selected from the examples for which there are at least nineteen non-empty constraint sets (since we set $t=2$ for CHI3D).}
    \label{fig:supp-chi3d-part2}
\end{figure*}

\begin{figure*}
    \centering
    \begin{sideways}
        \includegraphics[scale=0.28]{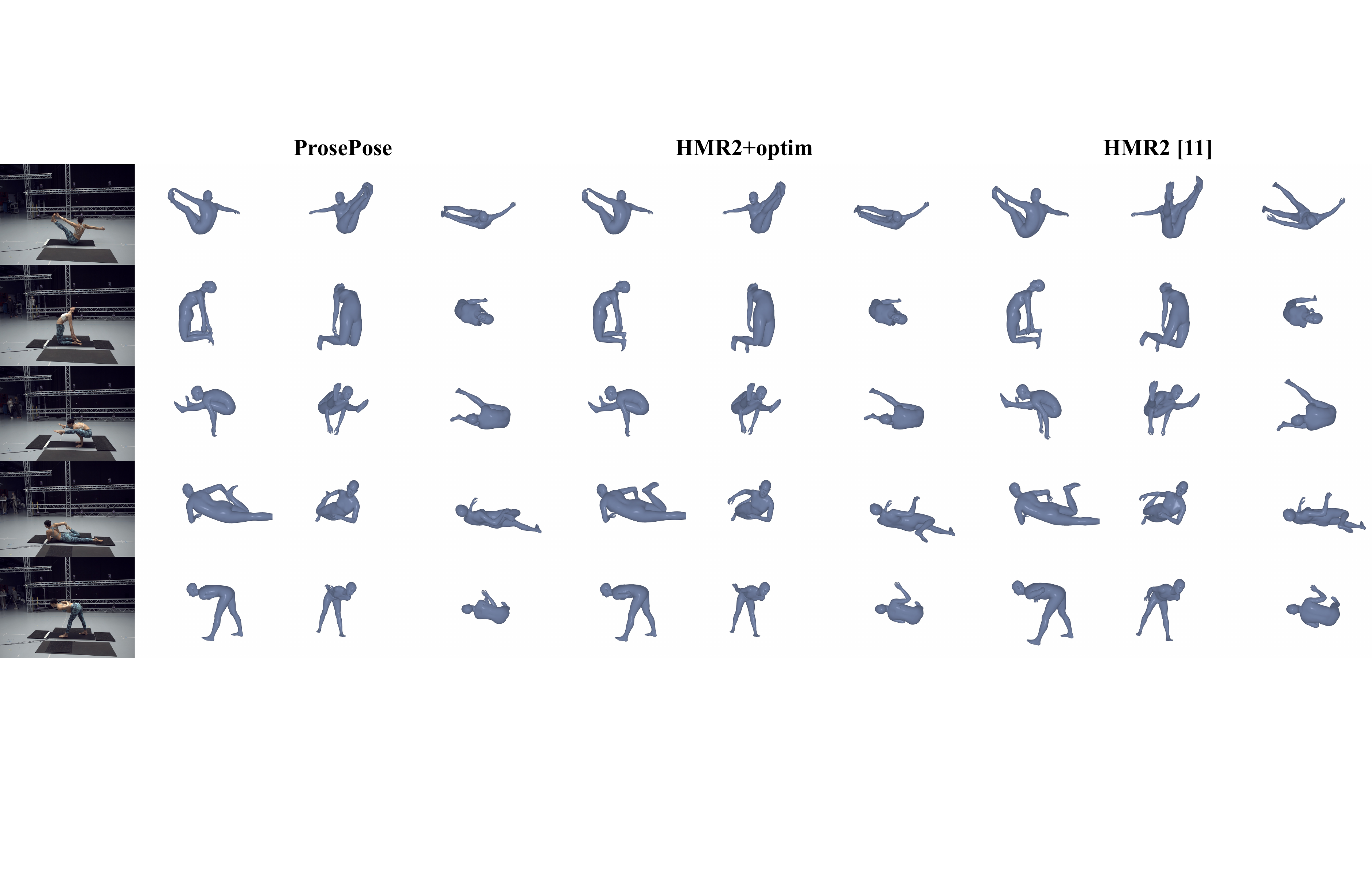}
    \end{sideways}
    \caption{Non-curated examples from the MOYO test set. They are randomly selected from the examples for which there is at least one non-empty constraint set.}
    \label{fig:supp-one-person-part1}
\end{figure*}

\begin{figure*}
    \centering
    \begin{sideways}
        \includegraphics[scale=0.28]{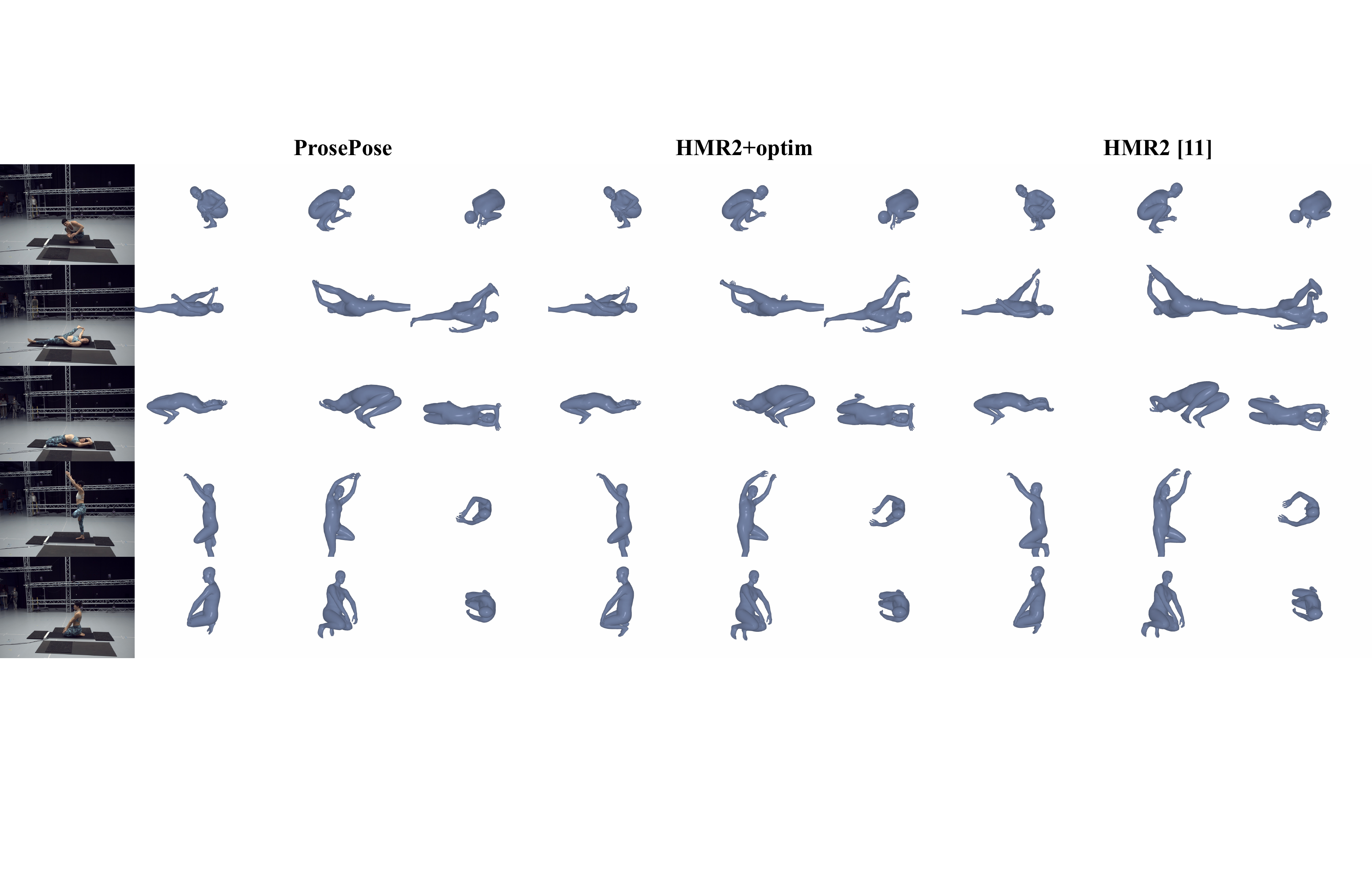}
    \end{sideways}
    \caption{Non-curated examples from the MOYO test set. They are randomly selected from the examples for which there is at least one non-empty constraint set.}
    \label{fig:supp-one-person-part2}
\end{figure*}

Figures~\ref{fig:supp-two-person-part1}, \ref{fig:supp-two-person-part2}, \ref{fig:supp-two-person-part3}, and \ref{fig:supp-two-person-part4} show additional, randomly selected examples from the multi-person FlickrCI3D test set. Figures~\ref{fig:supp-pgt-part1}, \ref{fig:supp-pgt-part2}, \ref{fig:supp-pgt-part3}, and \ref{fig:supp-pgt-part4} show the same examples comparing \methodname with the pseudo-ground truth fits. Figures~\ref{fig:supp-hi4d-part1}, \ref{fig:supp-hi4d-part2}, and \ref{fig:supp-hi4d-part3} show additional, randomly selected examples from the Hi4D test set. Figures~\ref{fig:supp-chi3d-part1} and \ref{fig:supp-chi3d-part2} show additional, randomly selected examples from the CHI3D validation set (which we use as the test set following \cite{muller2023generative}). Figures~\ref{fig:supp-one-person-part1} and \ref{fig:supp-one-person-part2} show additional, randomly selected examples from the 1-person yoga MOYO test set.
